# A Stroke-Level Large-Scale Database of Chinese Character Handwriting and the OpenHandWrite_Toolbox for Handwriting Research

Zebo Xu[1], Shaoyun Yu[1, 2], Mark Torrance[3], Guido Nottbusch[4], Nan Zhao[5], Zhenguang Cai[1,2]

1 Department of Linguistics and Modern Languages, The Chinese University of Hong Kong, Hong Kong SAR

2 Brain and Mind Institute, The Chinese University of Hong Kong, Hong Kong SAR

3 Psychology Department, School of Social Sciences, Nottingham Trent University, UK

4 Primary School Education, Human Science Faculty, University of Potsdam, Potsdam, Germany

5 Department of Translation, Interpreting and Intercultural Studies, Hong Kong Baptist University, Hong Kong SAR

**Author notes**

The research was supported by a GRF grant (14613722) from Research Grants Council (Hong Kong). Correspondence should be addressed to Z. G. Cai, Leung Kau Kui Building, The Chinese University of Hong Kong, Shatin, Hong Kong; zhenguangcai@cuhk.edu.hk.

**Abstract**

Understanding what linguistic components (e.g., phonological, semantic, and orthographic systems) modulate Chinese handwriting at the character, radical, and stroke levels remains an important yet understudied topic. Additionally, there is a lack of comprehensive tools for capturing and batch-processing fine-grained handwriting data. To address these issues, we constructed a large-scale handwriting database in which 42 Chinese speakers for each handwriting 1200 characters in a handwriting-to-dictation task. Additionally, we enhanced the existing handwriting package and provided comprehensive documentation for the upgraded *OpenHandWrite_Toolbox*, which can easily modify the experimental design, capture the stroke-level handwriting trajectory, and batch-process handwriting measurements (e.g., latency, duration, and pen-pressure). In analysing our large-scale database, multiple regression results show that orthographic predictors impact handwriting preparation and execution across character, radical, and stroke levels. Phonological factors also influence execution at all three levels. Importantly, these lexical effects demonstrate hierarchical attenuation - they were most pronounced at the character level, followed by the radical, and were weakest at the stroke levels. These findings demonstrate that handwriting preparation and execution at the radical and stroke levels are closely intertwined with linguistic components. This database and toolbox offer valuable resources for future psycholinguistic and neurolinguistic research on the handwriting of characters and sub-characters across different languages.

**INTRODUCTION**

In the past decades, researchers have made extensive use of digital tablets for collecting online handwriting data. These studies have demonstrated that handwriting involves multiple cognitive stages, from the preparation stages, such as lexical selection and orthographic access, as well as the execution stages, which involve transcoding phonological and semantic information into orthographic output (Wang et al., 2020; Damian & Qu, 2019; Zhang & Wang, 2015). In visually complex writing systems like Chinese, these processes may coordinate across multiple orthographic levels, since characters can decompose into radicals, and radicals into strokes. However, most handwriting research has focused on word-level processing, leaving sub-word handwriting features understudied. To investigate how phonological, semantic, and orthographic factors modulate handwriting performance across three orthographic levels, we developed a large-scale muti-level handwriting database and upgraded an open-source toolbox for the collection and analysis of online handwriting data.

*Chinese Character Handwriting in the Digital Age*

Unlike words in alphabetic script, Chinese characters are organised in a "square layout", which is the smallest unit that can function independently as a word with meaning (e.g., 清, *qing$_1$*, meaning "clear") or combine with other characters to form multi-character words (e.g., 清水, *qing$_1$shui$_3$*, meaning "clear water"). The Chinese writing system is renowned for its intricate hierarchy of orthographic levels: characters, radicals, and strokes. A character contains one or more radicals (e.g., 氵 and 青 in 清), which are composed of strokes (e.g., radical 氵 comprises strokes 、, 、, and 丿), arranged spatially within the square layout. In the character 清, which follows a left-to-right composition, 氵 appears on the left and 青 on the right. While the radical 青 (*qing$_1$*, meaning "green") can function independently as a character, the radical

氵 cannot. Learners must follow strict stroke order conventions when writing radicals and strokes to form characters. For instance, in 清, the radical 氵 is written first, followed by 青; within 氵, the stroke order is 丶, 丶, and ㇀.

Chinese characters are also known for their limited transparency between phonology and orthography, as phonemes do not map directly onto graphemes. For instance, the character 清 is pronounced as *qing₁*, (or /tɕʰiŋ/ in IPA), yet its phonemes (/tɕʰ/, /i/, or /ŋ/) do not individually corresponding to specific graphemes. Although some phonetic radicals can offer hints about pronunciation (e.g., 青 provides cues to the pronunciation of the character 清, *qing₁*), these cues are generally unreliable: the pronunciation of a character aligns with that of its phonetic radical only about 30% of the time, according to the Xinhua Dictionary and primary Chinese textbooks from grades 1 to 6 (Shu, 1998; Zhou, 1978). Additionally, the pronunciation of a single phonetic radical may correspond to various characters (e.g., 轻, 卿, 倾, these characters are all pronounced as "*qing₁*" but use different radicals), and the position of a phonetic radical is not consistent (e.g., the radical 青 appears in different positions in characters like 清, 静, and 箐). This lack of correspondence between phonemes and graphemes creates challenges for phonology-based typing systems, which dominate in mainland China. The digital age has consequently witnessed increasing difficulties for Chinese speakers in handwriting characters, a phenomenon termed "character amnesia" (提笔忘字 in Chinese; Almog, 2019; Huang et al., 2021a, 2021b), whereby individuals fail to handwrite characters they can readily recognize. Research on Chinese character handwriting therefore provides crucial insights into the cognitive processes underlying this unique writing system and may inform strategies to combat character amnesia (Duan & Cai, 2024; Xu & Cai, 2023). Understanding the preparation and execution stages of Chinese character handwriting enables researchers to identify potential intervention points for maintaining and improving handwriting skills in the digital era.

*Empirical studies on Chinese character handwriting*

In a large-scale study by Wang et al. (2020), 203 Chinese speakers participated in a handwriting-to-dictation task, in which they handwrote 200 characters randomly selected from a 1600-character pool. Multiple regression analyses revealed that handwriting preparation (i.e., as measured by retrieval accuracy and writing latency) was more accurate and quicker for characters with greater regularity, higher homophone density, imageability, concreteness, frequency, earlier age of acquisition, fewer strokes, and greater context word familiarity. These lexical effects also extended to writing execution, as decreased writing durations were associated with characters with higher frequency, earlier age of acquisition, fewer strokes, or greater word context familiarity. These results suggest that handwriting latency is a central process in the access and transcode of phonological, semantic, and orthographic information. Moreover, writing execution appears to interact with these central processes, with lexical effects cascading from writing preparation to execution (Baxter & Warrington, 1986; van Galen, 1991; Zhang & Feng, 2017).

There have been also attempts to investigate how lexical variables influence sub-character handwriting (i.e., radical and stroke level writing latency). For instance, Lau (2020a) recruited 20 native Cantonese-speaking adults, each copying 211 traditional Chinese characters. The regression analyses revealed that radical writing latency (i.e., the duration between finishing writing the last radical and the pen touching the tablet to start the next radical) was longer than stroke writing latency (i.e., duration between finishing writing the last stroke and the pen touching the tablet to start the next stroke) even after controlling for inter-stroke distance. Moreover, lower character frequency and a greater number of character strokes significantly increased radical and stroke writing latencies. These effects were more pronounced at the radical level than at the stroke level. Lau (2020b) replicated these findings and further showed that higher character regularity predicted longer radical and stroke writing latencies. Similar lexical effects on sub-lexical handwriting have also been observed in adults learning Chinese as a foreign language (Zhang, 2022). These results suggest that radical and stroke level writing preparation (i.e., writing latency) is modulated by phonological and orthographic processes, implying that lexical effects cascade from character to radical and stroke level handwriting latency.

However, existing research on sub-character level handwriting mainly used a small number of stimuli, which might restrict the generalizability of the findings across the overall character set. This limitation is particularly problematic for Chinese, given its vast inventory of characters with complex internal structures and considerable variability in radical and stroke configurations. For instance, the 211 characters in Lau (2020a) represent less than 10% of the characters that literate Chinese speakers commonly encounter, potentially failing to capture the full range of orthographic, phonological, and semantic variations that exist in the Chinese writing system. The constraints of small-scale studies become even more pronounced when investigating sub-character components. With limited stimuli, it becomes difficult to disentangle the effects of various lexical variables that may be confounded at the radical and stroke levels. For instance, radicals vary not only in their positional constraints but also in their phonetic transparency, factors that cannot be adequately controlled or examined with small stimulus sets. Moreover, the interaction between character-level and sub-character-level variables requires sufficient statistical power to detect, which is challenging to achieve with restricted samples.

A large-scale approach to studying sub-character handwriting would address these limitations by examining a comprehensive range of characters that better represents the diversity of the Chinese writing system. By including a substantial number of characters with varying radical compositions, stroke patterns, and lexical properties, researchers can more reliably isolate the independent contributions of different variables while controlling for potential confounds. This approach would also enable the investigation of how lexical effects cascade from the character level to radical and stroke levels across different types of characters, providing a more complete picture of the cognitive processes underlying Chinese handwriting production. Furthermore, findings from large-scale studies would have greater ecological validity and generalizability, offering insights that are more representative of real-world Chinese handwriting behaviour. Indeed, recent large-scale studies have examined how word level processing interacts with lexical effects, including phonological (e.g., character regularity and homophone density), semantic (e.g., imageability and concreteness), and orthographic factors (e.g., character frequency and number of strokes), over recognition (Sze et al., 2014; Tse et al., 2017), naming (Chang et al., 2016), and handwriting (Wang et al., 2020). It has been also applied to study various languages, including English (e.g., Balota et al., 2007; Keuleers et al., 2012), French (e.g., Ferrand et al., 2010), Italian (Barca et al., 2002), and Chinese (e.g., Liu et al., 2007; Wang et al., 2020).

*Tools for capturing handwriting processes*

Existing handwriting research tools can capture various measurements, including character writing latency, duration, and pen-pressure (Li-Tsang et al., 2022). However, implementing flexible handwriting experiments tailored to diverse research purposes remains challenging. General-purpose platforms often lack user-friendly solutions and are constrained by platform capabilities and researchers' programming expertise. For example, Wang et al. (2020) used E-Prime for Chinese handwriting experiments but could not capture online measures like handwriting timecourse or pen-pressure due to E-Prime's lack of digitizer pen support, resulting in less precise measurements of writing trajectories and duration (e.g., handwriting offset was determined by participants pressing a keyboard). Specialized solutions offer more targeted functionality but come with their own limitations. The Smart Handwriting Analysis and Recognition Platform (SHARP; Li-Tsang et al., 2022) and Ductus (Guinet & Kandel, 2010) support basic experimental designs with text, picture, and audio presentations while capturing kinematic data including writing velocity, latency, and duration. Eye and Pen (Alamargot et al., 2006) enables synchronized collection of eye and pen movement data, facilitating research on relationships between handwriting duration and eye fixation duration (Drijbooms et al., 2020; Alamargot et al., 2010; Lambert et al., 2011; Alamargot et al., 2011). However, these specialized tools typically depend on proprietary services and dedicated hardware, limiting experimental design flexibility as they offer only basic templates and require advanced programming skills for customization, thus restricting broader accessibility. Additionally, the lack of standardized protocols for recording and storing handwriting data has impeded the research community's ability to share and compare findings across studies effectively.

Recent handwriting programs follow the Open Science Movement, allowing researchers to freely use, modify, and distribute the software. OpenHandWrite (Simpson et al., 2021) is a platform designed to address the above-mentioned challenges by facilitating experimental design, data preprocessing and the sharing of handwriting research. It integrates advanced capabilities for capturing detailed handwriting trajectories, such as zero-pressure hovering movements, with the versatility of PsychoPy (Peirce et al., 2019), a widely used open-source application for building experiments. OpenHandWrite also saves and exports experimental data in standardised formats, offering an open protocol for exchanging handwriting datasets among researchers.

*The Present Study*

Wang et al. (2020) employed large-scale experiments to investigate the cognitive processes underlying character-level handwriting. Lau (2020a, 2020b) further examined how phonological and orthographic processes (e.g., character regularity, character frequency, and stroke number) influence sub-character handwriting at both the radical and stroke levels. However, these studies neither provided accurate handwriting measurements nor conducted large-scale experiments that accounted for the influence of other lexical variables (e.g., imageability, concreteness, or age of acquisition). Existing tools for handwriting research are mainly proprietary products (e.g., Wang et al., 2020; Li-Tsang et al., 2022), which prohibit customization for specific research requirements. To address these gaps, we collected online handwriting measures for 1,200 simplified Chinese characters from 42 adult participants and conducted multiple regression analyses to examine how lexical variables influence handwriting at the character, radical, and stroke levels. We also developed *OpenHandWrite_Toolbox*, a user-friendly suite of tools built on top of OpenHandWrite (Simpson et al., 2021), which we used to collect online Chinese handwriting data. The toolbox includes a graphical user interface (GUI) for intuitive experiment design and companion batch-processing scripts to extract critical handwriting variables, including writing latency, writing duration, and pen pressure at the character, radical, and stroke levels. It also supports annotation of fine-grained writing trajectories and the digitization of handwritten images. Finally, this study provides a comprehensive and user-friendly guide to the full functionality of the toolbox, enabling a wider audience to use, adapt, and modify it for their research.

**TOOLBOX DEVELOPMENT**

OpenHandWrite (Simpson et al., 2021) is an open-source platform that captures detailed handwriting trajectories, including zero-pressure hovering movements, integrated with PsychoPy (Peirce et al., 2019), a widely used open-source application for building experiments. OpenHandWrite saves and exports experimental data in standardized formats, offering an open protocol for sharing handwriting datasets among researchers. We developed OpenHandWrite_Toolbox on the foundation of OpenHandWrite tools, which includes GetWrite and MarkWrite. GetWrite supports handwriting data collection using digitizer pens and Wacom tablets within the PsychoPy environment, while MarkWrite provides visualization and markup capabilities for segmenting handwriting traces into theoretically meaningful units.

The first major component of OpenHandWrite_Toolbox is a ready-to-use experimental template that researchers can easily customize using PsychoPy Builder. While previous studies have employed GetWrite in script-based PsychoPy experiments (Fitjar et al., 2021; 2024), our template integrates its functionality directly into Builder's graphical interface. Through Builder, researchers can visually construct experiments by connecting routines and loops. As illustrated in our experiment (**Figure 1**), a typical trial comprises several routines representing stages such as audio prompt presentation, handwriting recording, and self-report collection. Once an experiment is set up based on the experimental template, it can automatically record handwriting trajectories for each trial. A detailed guide to experiment building is provided in **Appendix A**.

The second key component of our package is a set of scripts designed to automatically compute and export handwriting metrics for analysis. We store handwriting data in the standard OpenHandWrite format, comprising timestamps, x- and y-coordinates, pen-pressure, and other experimental variables. This format ensures compatibility with MarkWrite, the data visualization and segmentation tools provided by OpenHandWrite, allowing researchers to inspect and group pen samples intuitively (**Figure 2**). For instance, multiple strokes can be grouped into radicals, which are essential components of Chinese characters. To extract meaningful features from raw data, our scripts automatically compute various metrics including writing latency, duration, pen trace length, and average pressure. These metrics are extractable at stroke, radical, and character levels (**Table 1**), supporting hierarchical analysis of handwriting production. The scripts also generate character images (**Figure 3**) and provide sophisticated visualizations that decompose characters into individual strokes, annotating each with associated metrics (**Figure 4**). Appendix B presents a detailed walkthrough of the scripts and metrics.

Taken together, OpenHandWrite_Toolbox delivers a complete pipeline for handwriting research, spanning experiment design, data collection, character segmentation, and feature extraction. This integrated package enables researchers to conduct fine-grained analyses of handwriting behavior and investigate its cognitive underpinnings.

Table 1 Description of the character, radical, and stroke level handwriting metrics.

|  | Description |
| --- | --- |
| Character writing latency | Duration from offset of the stimulus to the pen-tip firstly touched the tablet on this trial. |
| Character writing duration | Duration from the onset of pen-tip firstly touched the tablet on this trial to the last pen sample with non-zero pressure. |
| Character writing length | Sum of the lengths of all its strokes belonging to the character, it was calculated by the linear distance between consecutive pen sample points based on the x and y coordinates. |
| Character average pen-pressure | Mean pressure of all stroke samples of the character. |
| Radical writing latency | Duration between offset of the last radical to onset of the current radical. |
| Radical writing duration | Duration between onset of the current radical to the end of the current radical. |
| Radical writing length | Sum of the lengths of the strokes belonging to the radical. |
| Radical average pen-pressure | Mean pressure of the radical's pen samples. |
| Radical distance | Linear distance between first point of the current radical and last point of previous radical. |
| Stroke writing latency | Duration between offset of the last stroke to onset of the current stroke. |
| Stroke writing duration | Difference between the timestamps of the last and first stroke writing samples. |
| Stroke writing length | Sum of the lengths of the strokes. |
| Stroke average pen-pressure | Mean pressure of the pen samples in the stroke. |
| Stroke distance | Linear distance between the current stroke's first sample and the previous stroke's last sample. |

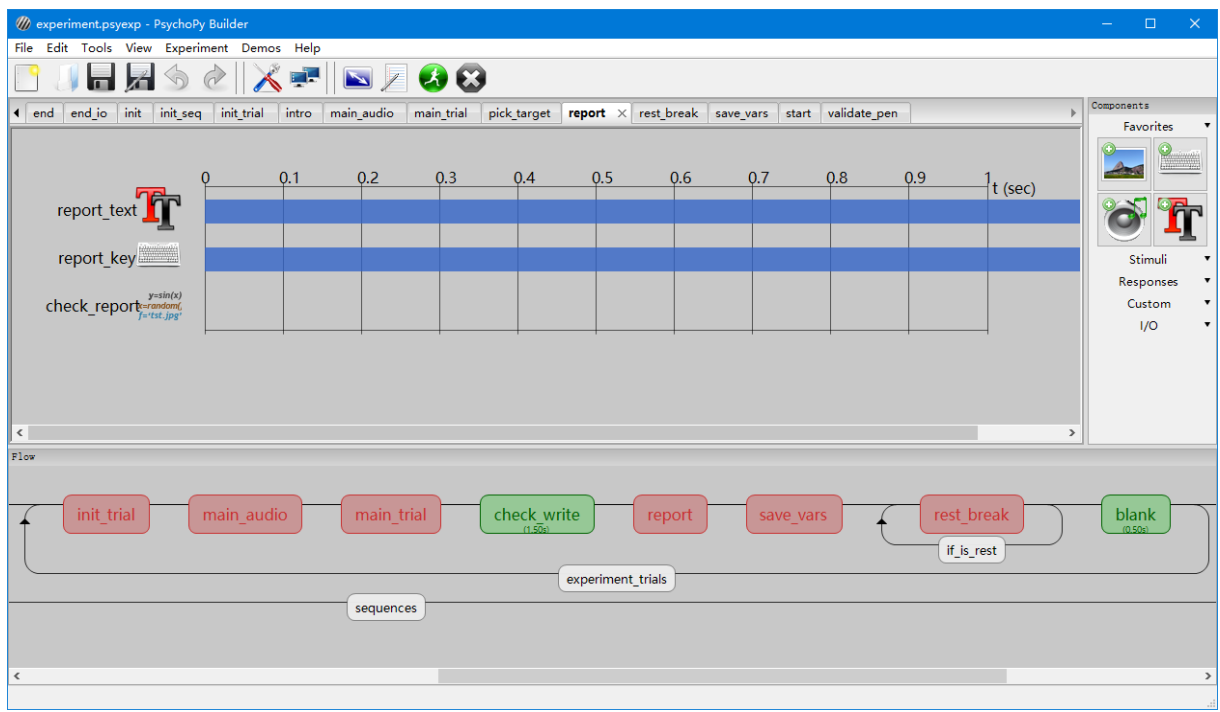

**Figure 1**. PsychoPy Builder interface. The lower panel displays the sequence of routines that define the experimental flow, with each routine represented as a coloured box. These routines are organised within a loop named "experiment_trials", which controls trial iteration. The upper panel shows the contents of the selected routine ("report"), where individual elements such as the instruction text ("report_text"), keyboard response ("report_key"), and custom code ("check_report") are arranged along a timeline. This visual interface allows users to build experiments without extensive programming.

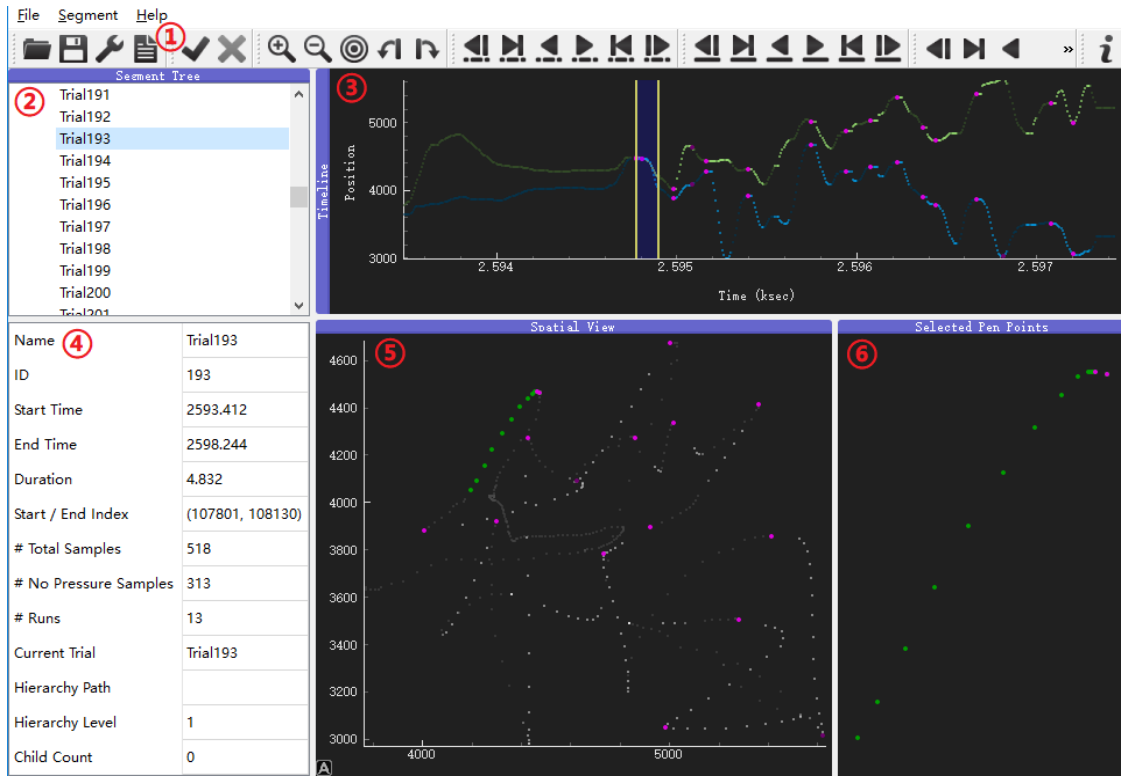

**Figure 2.** Markwrite interface. 1. Menu bar. 2. Segment Tree, which lists experimental trials. 3. Timeline, which displays pen points as a function of time (in milliseconds) and position (green points for horizontal positions and blue points for vertical positions). 4. Details of pen points selected in Timeline. 5. Spatial View, which displays the spatial view of the pen points, with the green dots representing the selected data point in Timeline, purple dots representing onset of a stroke, and gray dots representing pen in the air. 6. Selected Point Points, which is the zoom-in view of the data selected in Timeline.

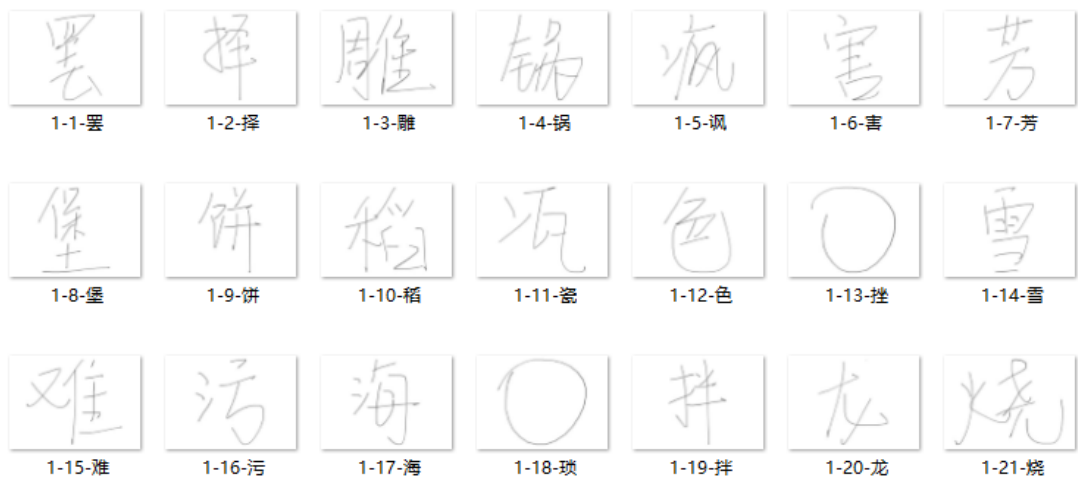

**Figure 3.** Examples of the handwritten images. For the file name of each image, the first number stands for the number of participants, the second number represents item number, following by the character name.

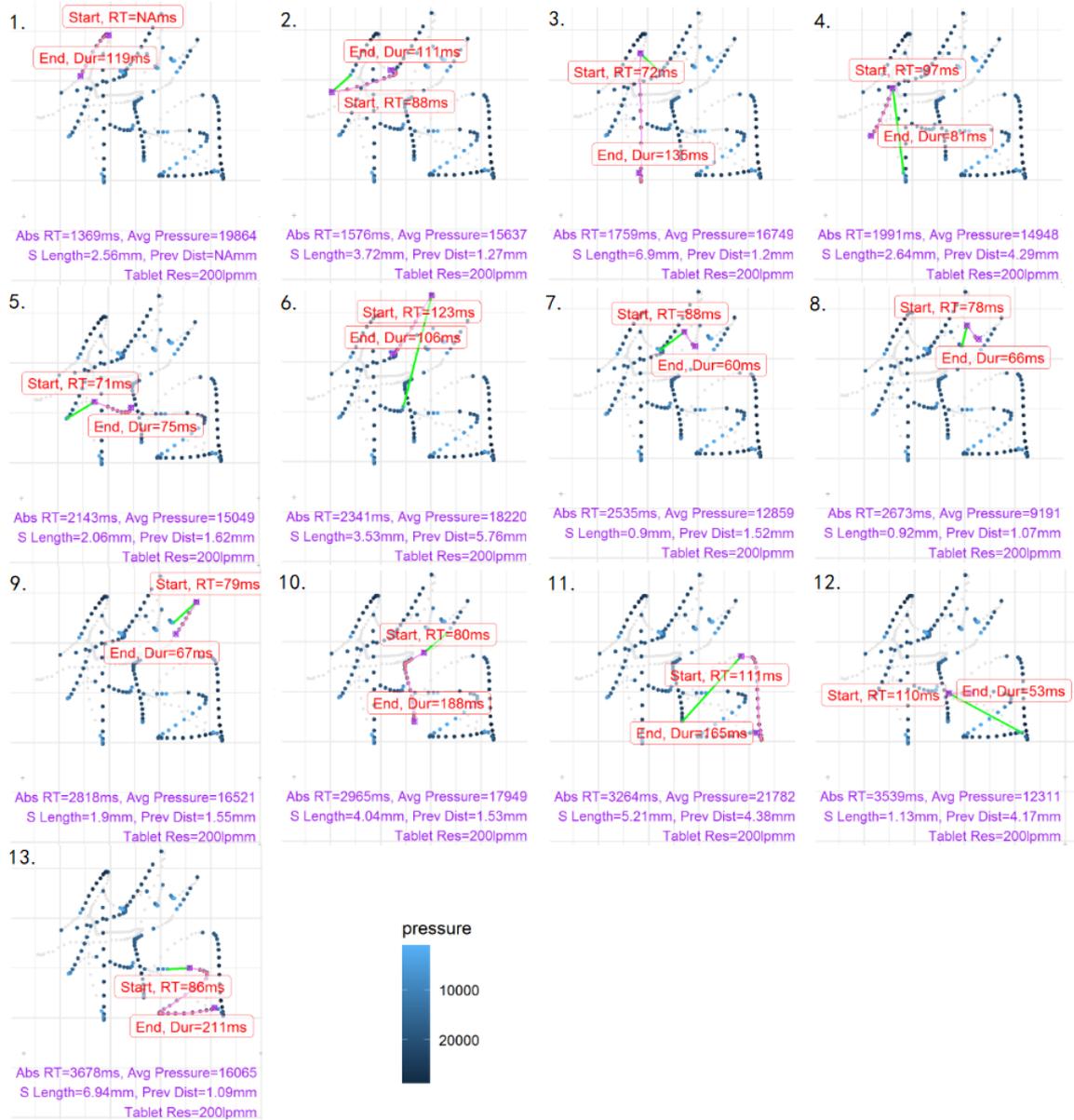

**Figure 4.** A plot showing the production for the handwritten character "稻" by one participant, with one panel for each of the 13 strokes, in the order in which the strokes were produced. The current stroke is represented by the purple line. The grey dots represent the pen in air movement. "Abs RT" refers to the duration from the offset of auditory stimuli to the start of the current stroke; "Avg Pressure" refers to the average pen-pressure for the current stroke; "S length" refers to the stroke length; "Prev Dist" refers to the distance between the start point of the current stroke and the previous stroke's end point, as represented by the green line. "Start, RT" refers to the stroke writing latency. "End, Dur" refers to the stroke writing duration.

**DEVELOPMENT OF A MULTI-LEVEL DATABASE OF CHINESE CHARACTER HANDWRITING**

We employed the OpenHandWrite_Toolbox to collect handwriting data through a handwriting-to-dictation task. Participants used an inking digitizer pen to write target characters as quickly and accurately as possible following spoken prompts. After writing, the target character appeared on screen, prompting participants to self-report whether their response was correct, a character amnesia response, or a "don't know" response. OpenHandWrite_Toolbox was used to present stimuli and record a multitude of handwriting data.

*Participants.* Forty-two native adult Mandarin speakers were recruited in this study (32 females; mean age = 20.50 years; range = 19-22 years). All were right-handed according to the Edinburgh Handedness Questionnaire (Oldfield, 1971), with normal hearing and normal or corrected-to-normal vision. None reported neurological or psychiatric disorders. All participants gave their written consent before participation. The protocol was approved by an ethics committee of The Chinese University of Hong Kong.

*Materials.* Characters were selected from the Chinese film subtitle corpus (SUBTLEX-CH; Cai & Brysbaert, 2010). As one of the purposes was to examine character amnesia in Chinese handwriting, we applied several filters to select a total of 1200 characters. Selection criteria included log frequency between 1.5 and 5.0 (corresponding to 32-100,000 corpus occurrences) and more than four strokes to avoid ceiling effects. These filters resulted in a candidate set of 2095 characters. For each character, we extracted the most common bi-character word from the SUBTLEX-CH (e.g., 水稻, shui$_3$ dao$_4$, for target character 稻, dao4). These words were then rated for familiarity by 15 participants on a scale from 1 to7 (the highest value represents the most familiar). These context words were included only when the familiar scores were equal or higher than 4 on average, resulting in 1600 candidate characters. These characters were then compiled into dictation phrases (e.g., 水稻的稻, the target character 稻, meaning "rice", in the word 水稻, meaning "wetland rice"). In order to reduce total time for the"). To reduce total time for the task, we randomly selected 1200 items from this candidate pool. The audios of the dictation phrases were then generated using Google Text-to-Speech (https://cloud.google.com/text-to-speech).

***Lexical variables.*** All the lexical variables for the selected characters were described by Wang et al. (2020). Each character was characterized by 14 phonological, semantic, and orthographic factors. We used the following four phonological variables. *Phonograms* refers to whether a character is a phonogram according to the Dictionary of Modern Chinese Phonograms (现代汉字形声字字汇; Ni, 1982). *Sound radical order* refers to whether a character with a sound radical to be written first (coded as 1; non-phonograms were categorised as not having the sound radical first). *Character regularity* refers to the extent to which the character contained a sound radical that indicated the pronunciation of the character. *Homophone* density refers to the number of characters that have the same pronunciation with a target character. There were three semantic variables. *Number* of meanings refers to the number of meanings a character has according to the newest Xinhua Dictionary (11th edition, Linguistics Institute of the Chinese Academy of Social Sciences, 2011). *Imageability* and *concreteness* refers to the perceived imageability and concreteness of a character as determined by subjective rating. Finally, we also considered the following five orthographic variables. *Character* frequency refers to the log of a character's count in the Chinese subtitle corpus SUBTLEX-CH (Cai & Brysbaert, 2010). *Age* of acquisition refers to the age a child learns a character according to the objective age of acquisition of Chinese characters in Cai et al. (2021) and Shu et al. (2003). *Number of strokes* refers to the number of strokes in a character according to Dictionary of Common Chinese Characters in Print (印刷通用汉字字形表; Chinese Ministry of Culture & State Language Affairs Commission, 1986) and the Modern Dictionary of Common Characters in Chinese (现代汉语通用字表; Chinese Ministry of Culture & State Language Affairs Commission, 1988). *Number of radicals* refers to the number of radicals in a character according to the Dictionary of Chinese Character Properties (汉字属性字典; Fu, 1989). *Character* composition refers to the organization of radicals within a character, according to the *Dictionary of Chinese Character Properties*, dummy-coded into two variables: whether a character has a left–right composition (coded as 1) or not (coded as 0), and whether a character has a top-down composition (coded as 0) or not (coded as 1). *Context* word familiarity refers to the familiarity of the context words.

***Apparatus and procedure.*** We adopted a handwriting-to-dictation task (**Figure 4, upper panel**) where participants first heard a cue sound, signalling the start of the experimental trial, followed by a 500ms blank interval. At this stage, participants were instructed to hold the pen above the correct grid square. They then heard a dictation prompt specifying a character to be handwritten (e.g., 水稻的稻, *shui$_3$ dao$_4$ de dao4*, the target character "稻" in the word "水稻"). Participants used an inking digitizer pen (Wacom KP-130-00DB) to write the target character in a grid on an 8 × 5 grid paper sheet affixed to a Wacom Intuos tablet (Wacom PTH-651; sampling rate set at 200 Hz), as quickly and accurately as possible after the offset of the audio; in case they did not know how to write the character, they were instructed to draw a circle in the grid. They then pressed the spacebar on the keyboard to indicate the completion of handwriting, which triggered the presentation of the target character for 1500 ms. Participants used the pen tip to press a key to self-report whether they had written the correct character (a correct handwriting response, number key 0), whether they knew which character was supposed to write but had forgotten how to write it (a character amnesia response, number key 1), or whether they did not know which character was supposed to write (a don't know response, number key 2), following by a 500 ms blank interval. Each participant took part in three experimental sessions, the interval among which was chosen by participants (which lasted from 2 hours to 4 days). In each session, participants handwrote 400 characters, with 40 characters on each 8 × 5 grid paper sheet from left to right, top to bottom (**Figure 5, lower** **panel**). The experiment automatically paused after the dictation of 40 characters, allowing the experimenter to replace the paper sheet.

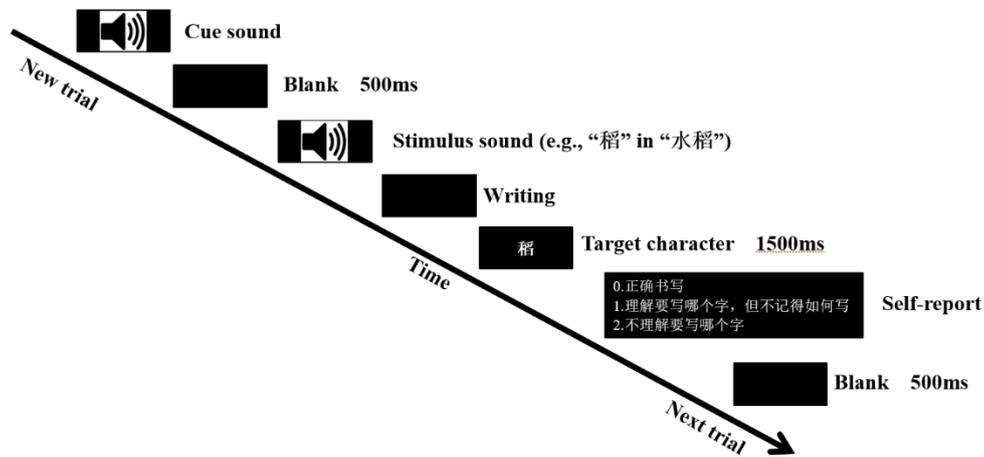

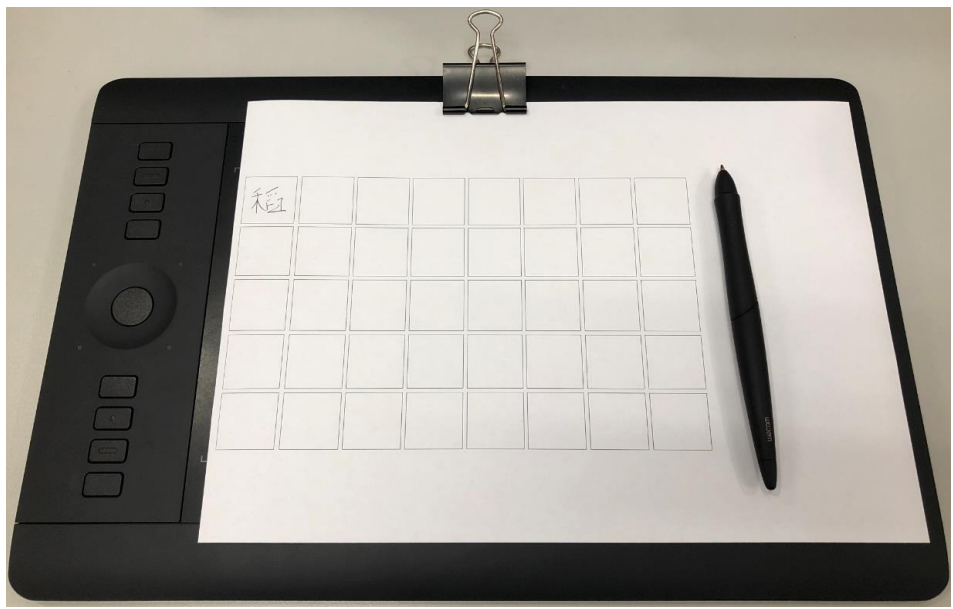

**Figure 5. Upper panel:** An example trial of the handwriting-to-dictation task. On a trial, participants heard a cue sound and, 500 ms later, a dictation prompt specifying a target character. They handwrote the target character as soon as possible. After participants finished handwriting, they were shown the target character on screen and self-reported whether they had a correct response, a character amnesia response, or a don't know response. There was a 500 ms inter-trial interval with a blank screen. **Lower panel:** An example of the 8 × 5 grid paper sheet on top of the tablet.

***Data coding and exclusion.*** Three helpers manually checked the penscripts (i.e., the collected handwritten images) against participants' self-reports on the online survey platform

Qualtrics (https://www.qualtrics.com/). The helpers were first shown a target character in its typed format, followed by the 42 corresponding penscripts of the character (from 42 participants). Each of them checked for 16800 penscripts of 400 target characters. They were instructed to mark a penscript as "correct handwriting" if they can recognise it as the target character (i.e., 44065 responses; 87.4% of the trials) and "incorrect handwriting" otherwise (i.e., 4950 responses; 9.8% of the trials). Additionally, they were instructed to mark if a penscript had been revised (e.g., with a false start, with a crossed character or radical), with a total of 1385 penscripts (2.7%) eventually marked as "revised". Incorrect and revised penscripts were subsequently excluded in the analyses of handwriting latency, duration, and pen-pressure. Additionally, we also removed handwriting latencies and durations at different levels of the character, the radical and the stroke. At the character level, we removed any writing latencies longer than 10 seconds and writing durations shorter than 1 second or longer than 10 seconds, leading to the exclusion of 0.39% of the writing latency data and 3.33% of the writing duration data. At the radical level, we removed latencies and durations longer than 2 seconds, leading to the exclusion of 0.28% of the writing latency data and 0.03% of the writing duration data. At the stroke level, we removed latencies and durations longer than 2 seconds, leading to the exclusion of 0.07% of the writing latency data and 0.02% of the writing duration data. Following previous large-scale studies on Chinese character recognition (Tsang et al., 2018; Sze et al., 2015), naming (Chang et al., 2016; Liu et al., 2007), and handwriting (Wang et al., 2020), we conducted item-level analyses by collapsing all the handwriting responses across participants, using the mean handwriting performance (e.g., accuracy or speed) for each unique character across in the following analyses. Table 2 presents descriptive results for the lexical variables and handwriting measures and Figure 6 represents the correlations (Person's r) among the variables for the 1200 Chinese characters. All stimuli are available on the Open Science Framework (https://osf.io/rn2ck/?view_only=4e2fc80957314e53a2afd47a1db8f217).

***Analyses of the database.*** We next provide some analyses of the database to investigate how handwriting performance is influenced by character-level variables. As there were correlations among the variables (shown in Figure 6), we assessed collinearity issues using stepwise variance inflation factor (VIF) selection (fmsb R package). With VIF threshold of 5 (Wang et al., 2020), all variables remain below threshold, indicating no serious collinearity among these lexical variables. We conducted multiple regression models to examine how different handwriting measures (character amnesia, handwriting latency, and handwriting duration) are influenced by the lexical variables. In the analyses, all the lexical variables were z-transformed and simultaneously entered into the regression models to examine how handwriting performance was modulated by the lexical variables: phonogram, sound radical order, regularity, homophone density, number of meanings, imageability, concreteness, frequency, age of acquisition, number of strokes, number of radicals, left–right composition, top-down composition, and word familiarity.

**Table 2** Descriptive results for lexical variables, character, radical, and stroke level handwriting data.

| Variable | Mean | SD | Range |
| --- | --- | --- | --- |
| Phonogram | 0.75 | 0.43 | 0.00-1.00 |
| Sound radical order | 0.18 | 0.39 | 0.00-1.00 |
| Regularity | -0.01 | 0.85 | -1.30-1.72 |
| Homophone density | 0.90 | 0.35 | 0.00-1.72 |
| Number of meanings | 3.20 | 2.10 | 1.00-16.00 |
| Imageability | -0.01 | 0.60 | -1.91-1.37 |
| Concreteness | -0.02 | 0.69 | -1.69-1.56 |
| Frequency | 3.44 | 0.68 | 1.58-5.00 |
| Age of acquisition | 8.50 | 1.63 | 6.50-15.00 |
| Number of strokes | 9.72 | 2.96 | 5.00-21.00 |
| Number of radicals | 2.95 | 1.04 | 1.00-7.00 |
| Left-right | 0.58 | 0.49 | 0.00-1.00 |
| Top-bottom | 0.73 | 0.45 | 0.00-1.00 |
| Word familiarity | 0.33 | 0.33 | -1.19-1.21 |
| Character latency | 1031.80 | 485.15 | 297.95-3861.67 |
| Character duration | 2222.05 | 656.29 | 1134.86-5011.30 |
| Character pressure | 12549.66 | 856.97 | 8779.62-15818.14 |
| Character length | 40.61 | 8.48 | 19.61-68.40 |
| Character amnesia rate | 0.05 | 0.08 | 0.00-0.59 |
| Radical latency | 151.30 | 45.01 | 61.38-547.05 |
| Radical duration | 707.17 | 310.87 | 104.05-2331.60 |
| Radical pressure | 12856.35 | 1211.96 | 6405.25-18084.43 |
| Radical length | 15.44 | 5.89 | 4.03-41.85 |
| Radical distance | 3.83 | 1.28 | 0.78-8.83 |
| Stroke latency | 113.11 | 28.68 | 78.86-398.69 |
| Stroke duration | 147.93 | 28.07 | 88.01-314.02 |
| Stroke pressure | 12096.85 | 1009.73 | 8323.09-16189.78 |
| Stroke length | 5.14 | 1.17 | 2.29-10.88 |
| Stroke distance | 2.51 | 0.50 | 1.20-5.29 |

Homophone density refers to the log of the number of homophonous characters in SUBTLEX-CH. Regularity, imageability, concreteness, and familiarity (context word familiarity) were based on ratings that were individually normalized. Frequency refers to the log of character counts in SUBTLEX-CH. Handwriting data in milliseconds.

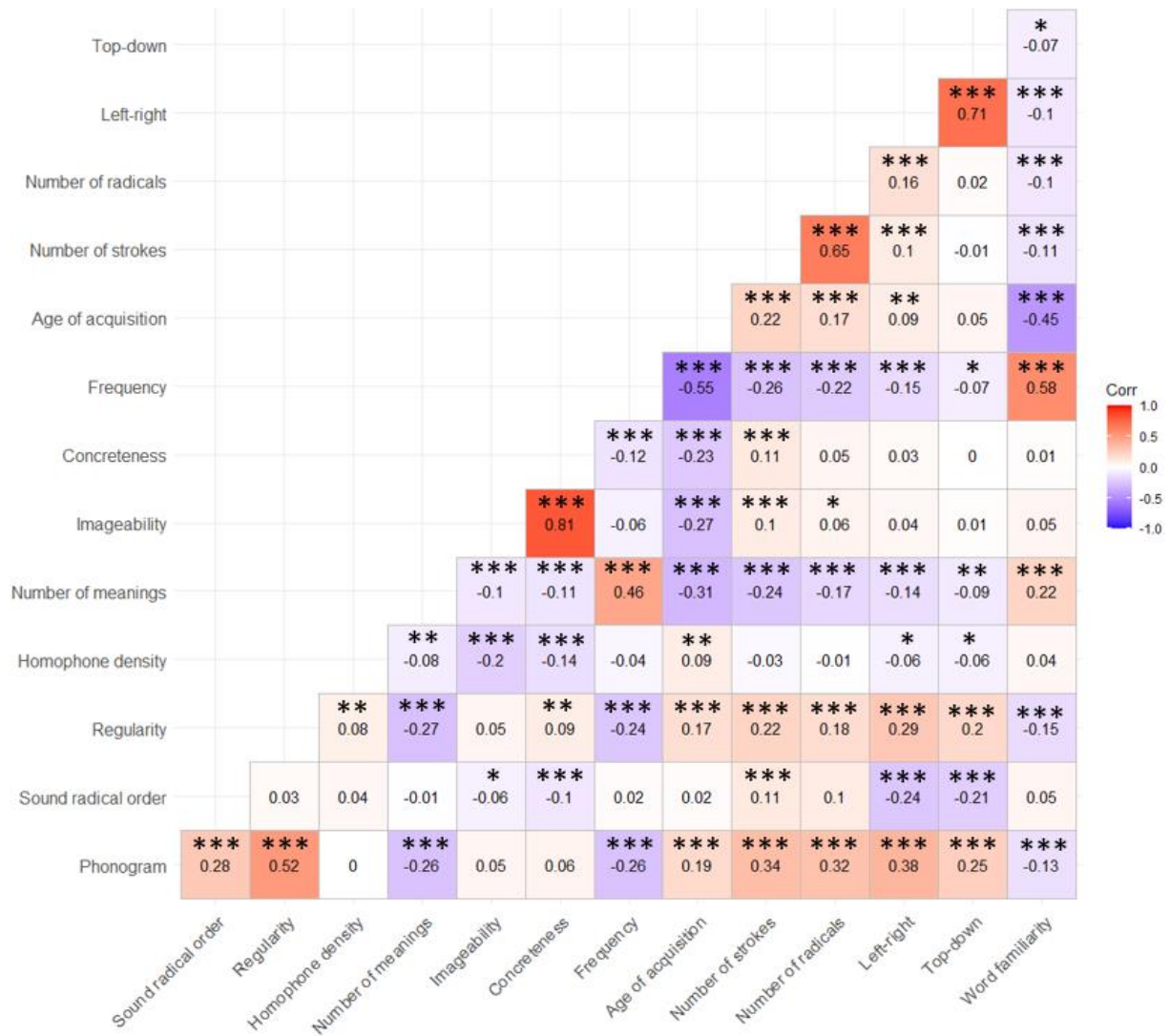

**Figure 6** Correlations of all variables. All the values listed within each matrix indicate correlation coefficients between variables (*** Correlation is significant at the .001 level; ** Correlation is significant at the .01 level; * Correlation is significant at the .05 level).

**Results**

**Character amnesia rate.** The results of the regression model ($R^2 = .426$) are presented in **Table 3**. The results show that phonology influences a character's tendency for amnesia: Characters are more likely to result in amnesia when the first radical is a sound radical than when it is not, or characters with lower regularity. Semantic predictors do not impact character amnesia rate. Orthographic factors modulate amnesia rate: Character amnesia rate is higher for characters that are less frequent, learned later, contain more strokes, being for a top-down composition, or appear in less familiar contextual words.

Table 3 Regressions on character amnesia rate.

| Linear term | Amnesia rate | | |
|---|---|---|---|
| | β | t | p |
| (Intercept) | 0.053 | 26.91 | **< .001\*\*\*** |
| Phonogram | -0.004 | -1.40 | 0.161 |
| Sound radical order | 0.006 | 2.85 | **0.005\*\*** |
| Regularity | -0.007 | -2.72 | **0.007\*\*** |
| Homophone density | -0.003 | -1.42 | 0.157 |
| Number of meanings | -0.001 | -0.41 | 0.683 |
| Imageability | -0.006 | -1.84 | 0.066 |
| Concreteness | 0.004 | 1.09 | 0.277 |
| Frequency | -0.022 | -7.59 | **< .001\*\*\*** |
| Age of acquisition | 0.028 | 10.70 | **< .001\*\*\*** |
| Number of strokes | 0.015 | 5.61 | **< .001\*\*\*** |
| Number of radicals | 0.002 | 0.63 | 0.532 |
| Left-right | -0.005 | -1.78 | 0.075 |
| Top-bottom | 0.008 | 2.84 | **0.005\*\*** |
| Word familiarity | -0.013 | -5.34 | **< .001\*\*\*** |

Significant p-values indicated in bold.
Key: β = coefficient, t = t-value, p = p-value.

**Character writing latency.** The results of the regression model ($R^2 = .488$) are presented in **Table 4**. Phonology modulates character writing preparation time: writing latency is shorter when the first radical is a sound radical than when it is not, or characters with lower homophone density. Semantic predictors impact writing preparation time, characters with higher concreteness predict shorter latency. Orthographically, characters with higher frequency, when learned at the earlier age, with fewer strokes, being for the left-right composition, not being for a top-bottom composition, or embedded in the more familiar words are associated with shorter writing latencies.

**Character writing duration.** The results of the regression model ($R^2 = .900$) are presented in **Table 4**. Phonology modulates character writing duration: writing duration is shorter when the character is a phonogram than when it is not. Orthography modulates character writing duration: it is shorter for characters with higher frequency, earlier age of acquisition, fewer strokes, or with left-right composition. Semantic predictors do not significantly affect writing duration.

**Character pen-pressure.** The results of the regression model ($R^2 = .601$) are presented in **Table 4**. Orthography modulates pen-pressure: pressure is greater for characters with higher frequency, earlier age of acquisition, fewer strokes, or left-right composition. Neither semantic nor phonological predictors significantly affect pen-pressure.

Table 4 Results of character level regressions on writing latency, writing duration, and character pen-pressure.

| Linear term | Character latency | | | Character duration | | | Character pen-pressure | | |
|---|---|---|---|---|---|---|---|---|---|
| | β | t | p | β | t | p | β | t | p |
| (Intercept) | 1031.16 | 102.33 | **<.001\*\*\*** | 1292.12 | 23.44 | **<.001\*\*\*** | 11886.32 | 83.23 | **<.001\*\*\*** |
| Phonogram | 0.62 | 0.04 | 0.965 | -19.21 | -2.26 | **0.024\*** | -20.77 | -0.94 | 0.346 |
| Sound radical order | 27.68 | 2.39 | **0.017\*** | 8.85 | 1.26 | 0.207 | -18.23 | -1.00 | 0.316 |
| Regularity | -3.45 | -0.28 | 0.777 | -5.09 | -0.69 | 0.489 | 16.76 | 0.88 | 0.379 |
| Homophone density | 25.76 | 2.46 | **0.014\*** | -1.16 | -0.18 | 0.854 | 9.57 | 0.58 | 0.559 |
| Number of meanings | 11.88 | 1.01 | 0.315 | -4.91 | -0.69 | 0.491 | 10.22 | 0.55 | 0.580 |
| Imageability | -12.70 | -0.72 | 0.469 | 14.44 | 1.37 | 0.172 | 46.02 | 1.68 | 0.093 |
| Concreteness | -48.26 | -2.75 | **0.006\*\*** | -6.10 | -0.58 | 0.564 | -37.81 | -1.38 | 0.168 |
| Frequency | -158.00 | -10.55 | **<.001\*\*\*** | -63.43 | -7.03 | **<.001\*\*\*** | 50.73 | 2.17 | **0.030\*** |
| Age of acquisition | 121.50 | 9.03 | **<.001\*\*\*** | 43.51 | 5.35 | **<.001\*\*\*** | -89.17 | -4.24 | **<.001\*\*\*** |
| Number of strokes | 51.34 | 3.69 | **<.001\*\*\*** | 423.13 | 33.41 | **<.001\*\*\*** | -710.69 | -21.66 | **<.001\*\*\*** |
| Number of radicals | -6.36 | -0.47 | 0.638 | 4.12 | 0.50 | 0.614 | -8.07 | -0.38 | 0.703 |
| Left-right | -36.51 | -2.32 | **0.020\*** | -51.07 | -5.38 | **<.001\*\*\*** | 120.88 | 4.92 | **<.001\*\*\*** |
| Top-bottom | 29.38 | 2.02 | **0.043\*** | 7.61 | 0.85 | 0.394 | -7.51 | -0.33 | 0.745 |
| Word familiarity | -105.34 | -8.26 | **<.001\*\*\*** | -13.22 | -1.72 | 0.086 | 20.95 | 1.05 | 0.293 |
| Character length | - | - | - | 22.90 | 16.97 | **<.001\*\*\*** | 16.32 | 4.67 | **<.001\*\*\*** |

Significant p-values indicated in bold.
Key: β = coefficient, t = t-value, p = p-value.

**Radical writing latency.** The results of the regression model ($R^2 = .291$) are presented in **Table 5**. Phonology influences radical writing preparation time: writing latency is shorter if a character's sound radical does not occur first. Orthography also modulates radical writing preparation time: A shorter radical writing latency is associated with characters that have higher frequency, are learned earlier, contain fewer strokes, have more radicals, or being for the left-right composition. Semantic predictors do not modulate writing preparation time.

**Radical writing duration.** The results of the regression model ($R^2 = .866$) are presented in **Table 5**. Phonology influences radical writing duration: duration is shorter when the character is not a phonogram or when its sound radical appears later in writing sequence. Orthographically, shorter radical writing duration associates with characters that have fewer strokes, have more radicals, being for the left-right composition, or not being for the top-down composition. Semantic predictors do not modulate writing duration.

**Radical pen-pressure.** The results of the regression model ($R^2 = .542$) are presented in **Table 5**. Phonology influences radical pen-pressure: pressure is greater when the characters are not phonograms, or their first radical are sound radicals. Semantic predictors also modulate pen-pressure, with higher pressure for more imageable characters. Orthographically, greater pen-pressure associates with characters contain fewer strokes, have fewer radicals, being for the left-right composition, or being for the top-down composition.

Table 5 Results of radical level regressions on writing latency, writing duration, and pen-pressure.

| Linear term | Radical latency | | | Radical duration | | | Radical pen-pressure | | |
|---|---|---|---|---|---|---|---|---|---|
| | β | t | p | β | t | p | β | t | p |
| (Intercept) | 112.84 | 27.99 | < .001*** | -10.05 | -0.61 | 0.541 | 13258.88 | 112.01 | < .001*** |
| Phonogram | -2.20 | -1.45 | 0.146 | 12.52 | 2.75 | **0.006**** | -80.10 | -2.44 | **0.015*** |
| Sound radical order | 5.52 | 4.17 | < .001*** | -14.63 | -3.68 | < .001*** | 90.48 | 3.16 | **0.002**** |
| Regularity | 0.30 | 0.23 | 0.822 | -4.28 | -1.06 | 0.289 | 16.49 | 0.57 | 0.570 |
| Homophone density | -1.53 | -1.30 | 0.193 | -1.51 | -0.43 | 0.669 | 2.72 | 0.11 | 0.915 |
| Number of meanings | -2.05 | -1.56 | 0.120 | 5.19 | 1.31 | 0.191 | 8.21 | 0.29 | 0.774 |
| Imageability | 2.89 | 1.49 | 0.138 | 3.28 | 0.56 | 0.575 | 99.68 | 2.37 | **0.018*** |
| Concreteness | 1.77 | 0.91 | 0.363 | -2.33 | -0.40 | 0.690 | -60.08 | -1.43 | 0.153 |
| Frequency | -7.41 | -4.48 | < .001*** | -8.96 | -1.80 | 0.072 | 2.69 | 0.08 | 0.940 |
| Age of acquisition | 7.50 | 5.08 | < .001*** | 7.07 | 1.59 | 0.112 | -53.82 | -1.69 | 0.092 |
| Number of strokes | 10.63 | 6.94 | < .001*** | 84.59 | 17.00 | < .001*** | -594.81 | -16.61 | < .001*** |
| Number of radicals | -5.87 | -3.87 | < .001*** | -21.90 | -3.71 | < .001*** | -341.88 | -8.05 | < .001*** |
| Left-right | -5.49 | -2.97 | **0.003**** | -40.47 | -7.24 | < .001*** | 413.07 | 10.27 | < .001*** |
| Top-bottom | 1.10 | 0.60 | 0.549 | 16.24 | 2.93 | **0.003**** | -206.60 | -5.18 | < .001*** |
| Word familiarity | -2.22 | -1.55 | 0.122 | 0.29 | 0.07 | 0.947 | 26.81 | 0.86 | 0.389 |
| Radical length | - | - | - | 46.89 | 59.69 | < .001*** | -6.76 | -1.20 | 0.232 |
| Radical distance | 10.04 | 9.94 | < .001*** | -1.78 | -0.59 | 0.559 | -77.83 | -3.55 | < .001*** |

Significant p-values indicated in bold.
Key: β = coefficient, t = t-value, p = p-value.

**Stroke writing latency.** The results of the regression model ($R^2 = .173$) are presented in **Table 6**. Orthography influences stroke writing latency: preparation time is shorter for characters that have higher frequency, are learned earlier, show left-right composition, being for top-down composition. Neither phonological nor semantic factors significantly affect stroke writing latency.

**Stroke writing duration.** The results of the regression model ($R^2 = .795$) are presented in **Table 6**. Phonology influences stroke duration: phonograms show shorter stroke execution than non-phonograms. Semantics also modulates duration, with shorter durations for characters exhibiting higher concreteness. Orthographically, shorter stroke execution associates with characters that contain fewer strokes, show left-right composition, lack top-down composition.

**Stroke pen-pressure.** The results of the regression model ($R^2 = .605$) are presented in **Table 6**. Orthography influences pen-pressure: greater stroke pen-pressure is associated with characters that are acquired earlier, contain fewer strokes, have left-right composition, being for a top-down composition, or appear in more familiar context words. Neither phonological nor semantic factors influence stroke pen-pressure.

**Table 6** Results of stroke level regressions on writing latency, writing duration, and pen-pressure.

| Linear term | Stroke latency | | | Stroke duration | | | Stroke pen-pressure | | |
| --- | --- | --- | --- | --- | --- | --- | --- | --- | --- |
| | β | t | p | β | t | p | β | t | p |
| (Intercept) | 73.41 | 17.08 | < .001*** | 50.13 | 16.67 | < .001*** | 10475.48 | 69.83 | < .001*** |
| Phonogram | 0.10 | 0.10 | 0.923 | -1.25 | -2.42 | **0.016*** | 4.21 | 0.16 | 0.870 |
| Sound radical order | -0.41 | -0.47 | 0.637 | 0.22 | 0.50 | 0.614 | -4.22 | -0.20 | 0.843 |
| Regularity | -0.75 | -0.82 | 0.413 | 0.64 | 1.44 | 0.151 | -3.50 | -0.16 | 0.876 |
| Homophone density | 0.74 | 0.94 | 0.350 | -0.45 | -1.17 | 0.244 | 13.55 | 0.71 | 0.481 |
| Number of meanings | 0.35 | 0.40 | 0.692 | -0.51 | -1.17 | 0.241 | 14.08 | 0.65 | 0.516 |
| Imageability | 0.46 | 0.35 | 0.726 | 1.12 | 1.74 | 0.082 | 19.85 | 0.62 | 0.537 |
| Concreteness | 1.36 | 1.03 | 0.302 | -1.40 | -2.17 | **0.030*** | -44.19 | -1.38 | 0.169 |
| Frequency | -3.47 | -3.08 | **0.002** | 0.15 | 0.27 | 0.788 | 52.06 | 1.90 | 0.058 |
| Age of acquisition | 3.30 | 3.26 | **0.001** | 0.11 | 0.22 | 0.825 | -58.90 | -2.39 | **0.017*** |
| Number of strokes | 0.63 | 0.60 | 0.552 | 5.70 | 10.27 | < .001*** | -486.27 | -17.55 | < .001*** |
| Number of radicals | 2.00 | 1.93 | 0.054 | 0.04 | 0.09 | 0.931 | 21.75 | 0.86 | 0.391 |
| Left-right | -11.02 | -9.31 | < .001*** | -1.40 | -2.42 | **0.016*** | 194.62 | 6.73 | < .001*** |
| Top-bottom | 8.59 | 7.83 | < .001*** | -1.45 | -2.59 | **0.010*** | -98.85 | -3.54 | < .001*** |
| Word familiarity | 0.91 | 0.95 | 0.344 | -0.88 | -1.87 | 0.062 | 66.53 | 2.85 | **005** |
| Stroke length | - | - | - | 23.55 | 61.37 | < .001*** | 337.76 | 17.64 | < .001*** |
| Stroke distance | 16.16 | 9.35 | < .001*** | -9.27 | -11.22 | < .001*** | -45.74 | -1.11 | 0.267 |

Significant p-values indicated in bold.
Key: β = coefficient, t = t-value, p = p-value.

**Hierarchical linguistic effects across different levels of writing latency.** Character latency is longer than both radical and stroke latencies, while radical latency is longer than stroke latency. Compared with writing latencies at both the radical and stroke levels, character latency has a greater increase for characters with more homophone density, lower concreteness, lower frequency, later age of acquisition, more strokes, or less word familiarity. Compared with stroke latency, character latency has a greater increase for characters with a sound radical to be written first. Compared with stroke latency, radical latency has a greater increase for characters with a sound radical to be written first, lower frequency, later age of acquisition, more strokes, fewer radicals or, or being for the left-right composition (**Table 7** and **Figure 7**).

**Table 7** Results of adult character, radical, and stroke level regressions on writing latency.

| Linear term | Level: Character vs. Radical | | | Level: Character vs. Stroke | | | Level: Radical vs. Stroke | | |
|---|---|---|---|---|---|---|---|---|---|
| | β | t | p | B | t | p | β | t | p |
| (Intercept) | 593.85 | 113.86 | **< .001*** | 572.13 | 113.2 | **< .001*** | 132.04 | 187.7 | **< .001*** |
| Level | 885.06 | 84.84 | **< .001*** | 918.05 | 90.83 | **< .001*** | 38.56 | 27.41 | **< .001*** |
| Phonogram | -1.17 | -0.16 | 0.870 | 0.28 | 0.04 | 0.969 | -1.50 | -1.56 | 0.120 |
| Sound radical order | 17.89 | 2.99 | **0.003** | 13.38 | 2.30 | **0.021*** | 3.46 | 4.28 | **< .001*** |
| Regularity | -1.25 | -0.20 | 0.841 | -2.26 | -0.37 | 0.712 | -0.06 | -0.08 | 0.940 |
| Homophone density | 12.16 | 2.25 | **0.025*** | 13.31 | 2.53 | **0.011*** | -0.27 | -0.37 | 0.712 |
| Number of meanings | 4.87 | 0.80 | 0.423 | 5.94 | 1.00 | 0.316 | -1.01 | -1.23 | 0.220 |
| Imageability | -5.01 | -0.56 | 0.578 | -6.15 | -0.70 | 0.485 | 1.49 | 1.23 | 0.220 |
| Concreteness | -23.45 | -2.61 | **0.009** | -23.50 | -2.67 | **0.008** | 1.11 | 0.91 | 0.362 |
| Frequency | -82.43 | -10.73 | **< .001*** | -80.67 | -10.7 | **< .001*** | -5.48 | -5.29 | **< .001*** |
| Age of acquisition | 64.51 | 9.31 | **< .001*** | 62.54 | 9.27 | **< .001*** | 5.49 | 5.88 | **< .001*** |
| Number of strokes | 32.08 | 4.54 | **< .001*** | 25.23 | 3.62 | **< .001*** | 6.26 | 6.57 | **< .001*** |
| Number of radicals | -8.54 | -1.25 | 0.212 | -3.16 | -0.47 | 0.642 | -5.44 | -5.89 | **< .001*** |
| Left-right | -21.10 | -2.54 | **0.011*** | -24.03 | -3.05 | **0.002** | -8.69 | -7.75 | **< .001*** |
| Top-bottom | 17.88 | 2.29 | **0.022*** | 19.50 | 2.68 | **0.007** | 7.99 | 7.58 | **< .001*** |
| Word familiarity | -53.95 | -8.19 | **< .001*** | -52.13 | -8.15 | **< .001*** | -0.62 | -0.70 | 0.485 |
| Level×Phonogram | 3.54 | 0.25 | 0.804 | 0.68 | 0.05 | 0.962 | -2.88 | -1.49 | 0.135 |
| Level×Sound radical order | 20.12 | 1.68 | 0.093 | 28.59 | 2.46 | **0.014*** | 8.75 | 5.42 | **< .001*** |
| Level×Regularity | -4.38 | -0.35 | 0.725 | -2.38 | -0.20 | 0.846 | 2.01 | 1.19 | 0.233 |
| Level×Homophone density | 27.13 | 2.51 | **0.012*** | 24.89 | 2.37 | **0.018*** | -2.28 | -1.56 | 0.119 |
| Level×Number of meanings | 13.78 | 1.13 | 0.257 | 11.87 | 1.00 | 0.317 | -2.03 | -1.24 | 0.216 |
| Level×Imageability | -15.18 | -0.84 | 0.399 | -13.11 | -0.75 | 0.456 | 2.18 | 0.90 | 0.370 |
| Level×Concreteness | -48.85 | -2.72 | **0.007** | -49.50 | -2.82 | **0.005** | -0.27 | -0.11 | 0.910 |
| Level×Frequency | -149.66 | -9.74 | **< .001*** | -154.60 | -10.3 | **< .001*** | -4.25 | -2.05 | **0.041*** |
| Level×Age of acquisition | 114.28 | 8.25 | **< .001*** | 117.86 | 8.73 | **< .001*** | 3.76 | 2.01 | **0.045*** |
| Level×Number of strokes | 37.41 | 2.65 | **0.008** | 52.20 | 3.74 | **< .001*** | 14.24 | 7.47 | **< .001*** |
| Level×Number of radicals | 4.76 | 0.35 | 0.728 | -6.42 | -0.47 | 0.637 | -10.97 | -5.95 | **< .001*** |
| Level×Left-right | -30.43 | -1.83 | 0.068 | -24.93 | -1.58 | 0.114 | 5.62 | 2.51 | **0.012*** |
| Level×Top-bottom | 23.12 | 1.48 | 0.139 | 19.74 | 1.36 | 0.175 | -3.33 | -1.58 | 0.114 |
| Level×Word familiarity | -103.29 | -7.84 | **< .001*** | -106.38 | -8.32 | **< .001*** | -3.37 | -1.90 | 0.058 |

Significant p-values indicated in bold.
Key: β = coefficient, t = t-value, p = p-value.

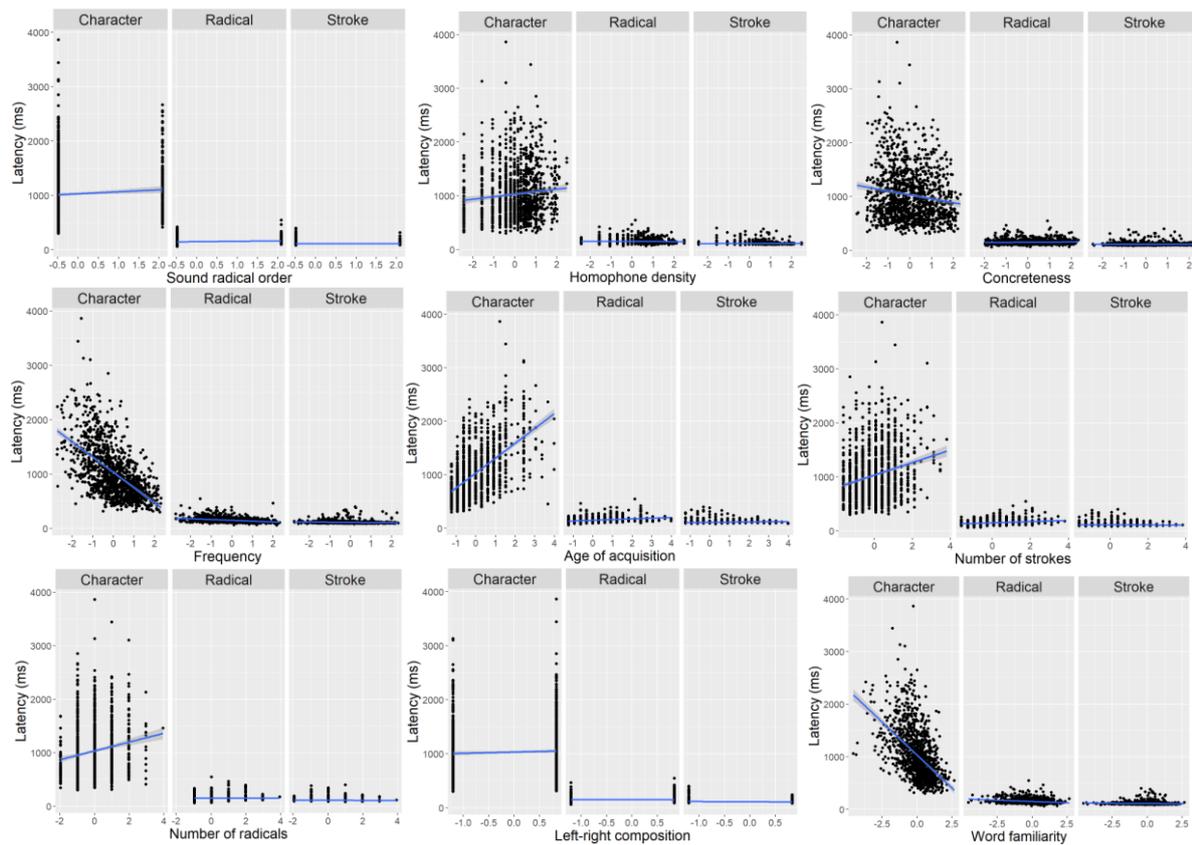

**Figure 7.** Significant interaction effects of writing latencies as a function of lexical variables, as described in **Table 7**. The lexical variables were transformed into z-scores, include sound radical order, homophone density, concreteness, frequency, age of acquisition, number of strokes, number of radicals, left-right composition, and word familiarity.

**Hierarchical linguistic effects across different levels of writing duration:** Character duration is longer than both radical and stroke durations, while stroke duration is longer than duration radical (after controlling for the length of writing). Compared with radical and stroke writing durations, character duration has a greater increase for characters that are not phonogram, have lower frequency, have later age of acquisition, contain more strokes. Compared with the radical writing duration, the increase of character writing duration is larger for characters with sound radicals to be written first or more radicals. Compared with the stroke duration, the increase of character duration is larger for characters not being for a left-right composition. Compared with stroke writing duration, radical duration has a larger increase for characters that are phonograms, with the sound radical to be written later, with more strokes, contain fewer radicals, not being for a left-right composition, or being for a top-bottom composition (**Table 8** and **Figure 8**).

Table 8 Results of character, radical, and stroke level regressions on writing duration.

| Linear term | Level: Character vs. Radical | | | Level: Character vs. Stroke | | | Level: Radical vs. Stroke | | |
|---|---|---|---|---|---|---|---|---|---|
| | β | t | p | B | t | p | B | t | p |
| (Intercept) | 644.07 | 26.43 | **< .001*** | 658.52 | 28.97 | **< .001*** | 4.04 | 0.53 | 0.599 |
| Level | 1322.32 | 27.13 | **< .001*** | 1267.19 | 27.87 | **< .001*** | -42.26 | -2.75 | **0.006** |
| Phonogram | -2.87 | -0.59 | 0.553 | -10.18 | -2.39 | **0.017*** | 5.95 | 2.60 | **0.009** |
| Sound radical order | -2.96 | -0.73 | 0.466 | 4.66 | 1.33 | 0.185 | -7.19 | -3.74 | **< .001*** |
| Regularity | -4.78 | -1.13 | 0.258 | -2.14 | -0.58 | 0.561 | -1.84 | -0.92 | 0.357 |
| Homophone density | -1.37 | -0.37 | 0.710 | -0.84 | -0.27 | 0.790 | -1.05 | -0.60 | 0.546 |
| Number of meanings | 0.22 | 0.05 | 0.958 | -2.61 | -0.73 | 0.465 | 2.49 | 1.28 | 0.201 |
| Imageability | 8.87 | 1.45 | 0.146 | 7.81 | 1.47 | 0.141 | 2.29 | 0.79 | 0.428 |
| Concreteness | -4.13 | -0.68 | 0.499 | -3.73 | -0.71 | 0.481 | -1.77 | -0.62 | 0.538 |
| Frequency | -36.00 | -6.91 | **< .001*** | -31.67 | -7.01 | **< .001*** | -4.39 | -1.78 | 0.075 |
| Age of acquisition | 25.40 | 5.40 | **< .001*** | 21.72 | 5.34 | **< .001*** | 3.59 | 1.62 | 0.106 |
| Number of strokes | 251.87 | 40.00 | **< .001*** | 214.91 | 39.07 | **< .001*** | 45.92 | 18.77 | **< .001*** |
| Number of radicals | -9.13 | -1.65 | 0.098 | 2.66 | 0.65 | 0.516 | -10.54 | -4.04 | **< .001*** |
| Left-right | -45.59 | -8.05 | **< .001*** | -26.05 | -5.47 | **< .001*** | -20.71 | -7.73 | **< .001*** |
| Top-bottom | 11.35 | 2.12 | **0.034*** | 2.70 | 0.60 | 0.551 | 6.43 | 2.52 | **0.012*** |
| Word familiarity | -6.45 | -1.44 | 0.150 | -7.10 | -1.85 | 0.065 | -0.32 | -0.15 | 0.881 |
| length | 34.87 | 43.66 | **< .001*** | 23.41 | 10.29 | **< .001*** | 35.39 | 28.68 | **< .001*** |
| Level×Phonogram | -31.78 | -3.28 | **0.001** | -18.05 | -2.12 | **0.034*** | 14.15 | 3.10 | **0.002** |
| Level×Sound radical order | 23.79 | 2.93 | **0.003** | 8.37 | 1.19 | 0.233 | -15.34 | -3.99 | **< .001*** |
| Level×Regularity | -0.59 | -0.07 | 0.945 | -5.90 | -0.80 | 0.424 | -5.29 | -1.32 | 0.186 |
| Level×Homophone density | 0.41 | 0.06 | 0.955 | -0.64 | -0.10 | 0.920 | -1.05 | -0.30 | 0.763 |
| Level×Number of meanings | -10.15 | -1.23 | 0.219 | -4.60 | -0.64 | 0.519 | 5.60 | 1.44 | 0.151 |
| Level×Imageability | 10.91 | 0.89 | 0.372 | 13.27 | 1.25 | 0.210 | 2.26 | 0.39 | 0.696 |
| Level×Concreteness | -3.86 | -0.32 | 0.752 | -4.74 | -0.45 | 0.654 | -0.85 | -0.15 | 0.883 |
| Level×Frequency | -54.27 | -5.21 | **< .001*** | -63.50 | -7.03 | **< .001*** | -8.95 | -1.82 | 0.069 |
| Level×Age of acquisition | 36.33 | 3.86 | **< .001*** | 43.56 | 5.36 | **< .001*** | 7.30 | 1.65 | 0.100 |
| Level×Number of strokes | 333.47 | 26.48 | **< .001*** | 416.27 | 37.84 | **< .001*** | 78.43 | 16.03 | **< .001*** |
| Level×Number of radicals | 26.23 | 2.38 | **0.018*** | 2.92 | 0.36 | 0.721 | -23.40 | -4.49 | **< .001*** |
| Level×Left-right | -10.41 | -0.92 | 0.358 | -50.02 | -5.26 | **< .001*** | -39.35 | -7.34 | **< .001*** |
| Level×Top-bottom | -7.45 | -0.70 | 0.486 | 9.82 | 1.09 | 0.278 | 17.29 | 3.39 | **< .001*** |
| Level×Word familiarity | -13.61 | -1.52 | 0.129 | -12.23 | -1.59 | 0.113 | 1.35 | 0.32 | 0.750 |
| Level× length | -23.95 | -15.00 | **< .001*** | -1.03 | -0.23 | 0.821 | 22.92 | 9.29 | **< .001*** |

Significant p-values indicated in bold.
Key: β = coefficient, t = t-value, p = p-value.

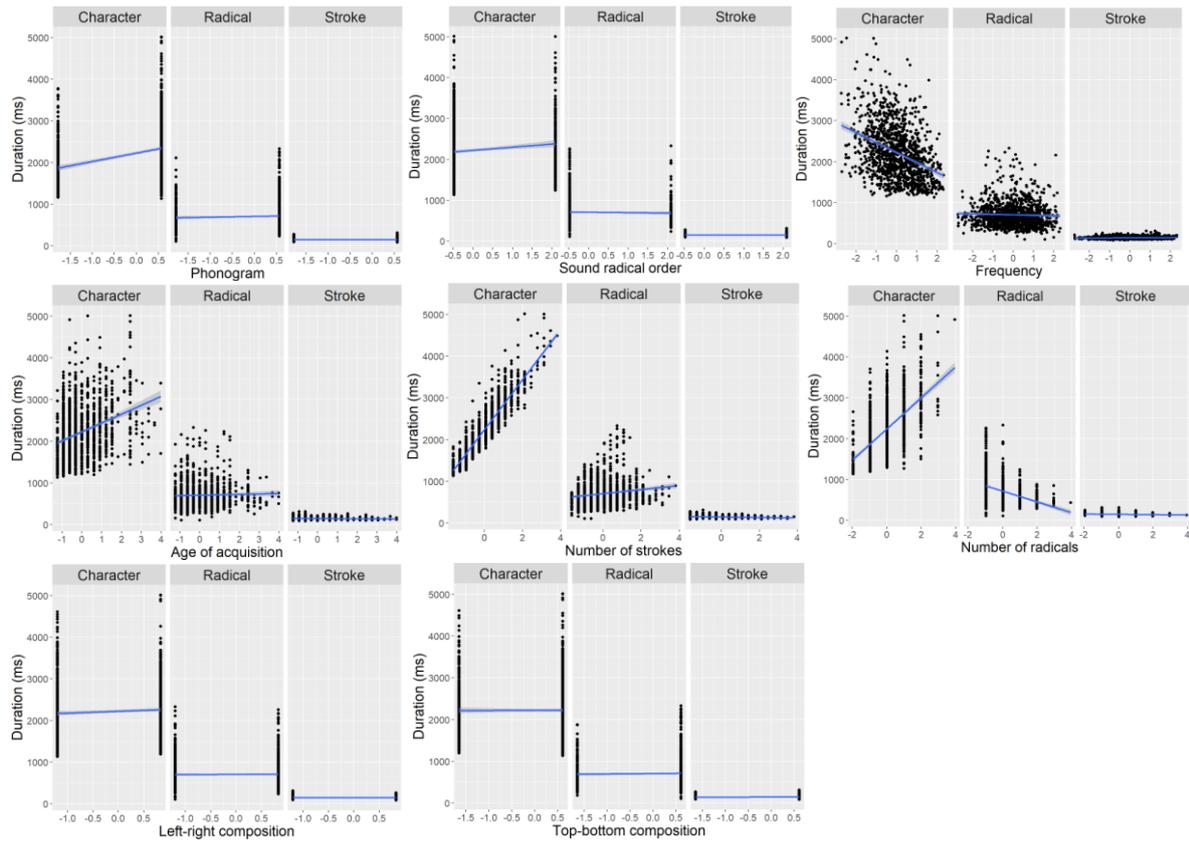

**Figure 8.** Significant interaction effects of writing durations as a function of lexical variables, as described in **Table 8**. The lexical variables were transformed into z-scores, include phonogram, sound radical order, frequency, age of acquisition, number of strokes, number of radicals, left-right composition, and top-bottom composition.

**Hierarchical linguistic effects at different levels of writing pen-pressure:** Radical pen-pressure is the largest, followed by character pen-pressure, and then stroke pen-pressure. Compared with both character and stroke pen-pressure, radical pen-pressure has a larger increase for characters with the sound radical to be written first, with fewer radicals, being a left-right composition. While radical pen-pressure has a larger decrease for characters being a top-bottom composition. Compared with stroke pen-pressure, character pen-pressure has a smaller increase for characters being for a left-right composition. Compared with stroke pen-pressure, character pen-pressure has a greater decrease for characters contain more strokes. Compared with stroke pen-pressure, character pen-pressure has a smaller decrease for characters have a top-bottom composition (**Table 9** and **Figure 9**).

**Table 9** Results of character, radical, and stroke level regressions on pen-pressure.

| Linear term | Level: Character vs. Radical | | | Level: Character vs. Stroke | | | Level: Radical vs. Stroke | | |
|---|---|---|---|---|---|---|---|---|---|
| | β | t | p | β | t | p | β | t | p |
| (Intercept) | 12439.0 | 124.70 | **< .001*** | 11118.7 | 122.75 | **< .001*** | 11677.5 | 165.92 | **< .001*** |
| Level | -1149.8 | -5.76 | **< .001*** | 1535.23 | 8.48 | **< .001*** | 2672.88 | 18.99 | **< .001*** |
| Phonogram | -48.55 | -2.45 | **0.014*** | -8.04 | -0.47 | 0.635 | -36.12 | -1.73 | 0.085 |
| Sound radical order | 26.36 | 1.58 | 0.113 | -10.58 | -0.76 | 0.449 | 34.07 | 1.94 | 0.053 |
| Regularity | 14.12 | 0.82 | 0.415 | 7.03 | 0.48 | 0.632 | 4.42 | 0.24 | 0.809 |
| Homophone density | 5.80 | 0.39 | 0.699 | 11.37 | 0.90 | 0.368 | 7.60 | 0.48 | 0.632 |
| Number of meanings | 8.96 | 0.53 | 0.596 | 12.64 | 0.89 | 0.374 | 11.35 | 0.64 | 0.525 |
| Imageability | 73.82 | 2.95 | **0.003**** | 33.05 | 1.57 | 0.117 | 60.96 | 2.31 | **0.021*** |
| Concreteness | -46.38 | -1.86 | 0.063 | -40.90 | -1.94 | 0.053 | -49.45 | -1.87 | 0.061 |
| Frequency | 25.79 | 1.21 | 0.227 | 51.22 | 2.85 | **0.004**** | 26.29 | 1.17 | 0.244 |
| Age of acquisition | -74.27 | -3.86 | **< .001*** | -74.44 | -4.59 | **< .001*** | -59.54 | -2.93 | **0.003**** |
| Number of strokes | -660.27 | -25.62 | **< .001*** | -595.70 | -27.18 | **< .001*** | -546.64 | -24.40 | **< .001*** |
| Number of radicals | -167.63 | -7.42 | **< .001*** | 9.69 | 0.59 | 0.552 | -150.45 | -6.30 | **< .001*** |
| Left-right | 269.98 | 11.64 | **< .001*** | 158.62 | 8.36 | **< .001*** | 307.55 | 12.53 | **< .001*** |
| Top-bottom | -126.99 | -5.81 | **< .001*** | -55.05 | -3.06 | **0.002**** | -174.66 | -7.47 | **< .001*** |
| Word familiarity | 24.01 | 1.31 | 0.190 | 43.45 | 2.83 | **0.005**** | 46.57 | 2.40 | **0.016*** |
| length | 3.92 | 1.20 | 0.231 | 177.97 | 19.64 | **< .001*** | 165.57 | 14.66 | **< .001*** |
| Level×Phonogram | 56.54 | 1.43 | 0.154 | -25.45 | -0.75 | 0.453 | -81.39 | -1.94 | 0.052 |
| Level×Sound radical order | -89.52 | -2.69 | **0.007**** | -15.28 | -0.55 | 0.585 | 74.09 | 2.11 | **0.035*** |
| Level×Regularity | 5.18 | 0.15 | 0.881 | 19.45 | 0.66 | 0.508 | 14.21 | 0.39 | 0.698 |
| Level×Homophone density | 7.50 | 0.25 | 0.803 | -3.61 | -0.14 | 0.886 | -11.10 | -0.35 | 0.727 |
| Level×Number of meanings | 2.33 | 0.07 | 0.945 | -4.84 | -0.17 | 0.865 | -7.12 | -0.20 | 0.842 |
| Level×Imageability | -56.32 | -1.13 | 0.260 | 25.93 | 0.61 | 0.539 | 82.05 | 1.55 | 0.121 |
| Level×Concreteness | 17.73 | 0.36 | 0.723 | 6.19 | 0.15 | 0.883 | -11.58 | -0.22 | 0.826 |
| Level×Frequency | 49.41 | 1.16 | 0.247 | -1.01 | -0.03 | 0.978 | -50.42 | -1.12 | 0.264 |
| Level×Age of acquisition | -30.03 | -0.78 | 0.435 | -29.42 | -0.91 | 0.364 | 0.56 | 0.01 | 0.989 |
| Level×Number of strokes | -85.64 | -1.66 | 0.097 | -229.68 | -5.24 | **< .001*** | -141.64 | -3.16 | **0.002**** |
| Level×Number of radicals | 319.65 | 7.08 | **< .001*** | -35.52 | -1.09 | 0.276 | -354.01 | -7.42 | **< .001*** |
| Level×Left-right | -299.49 | -6.46 | **< .001*** | -75.54 | -1.99 | **0.047*** | 224.34 | 4.57 | **< .001*** |
| Level×Top-bottom | 238.93 | 5.46 | **< .001*** | 95.09 | 2.64 | **0.008**** | -143.60 | -3.07 | **0.002**** |
| Level×Word familiarity | -6.02 | -0.16 | 0.870 | -45.01 | -1.47 | 0.143 | -39.11 | -1.01 | 0.313 |
| Level× length | 24.81 | 3.80 | **< .001*** | -323.30 | -17.83 | **< .001*** | -348.11 | -15.41 | **< .001*** |

Significant p-values indicated in bold.
Key: β = coefficient, t = t-value, p = p-value.

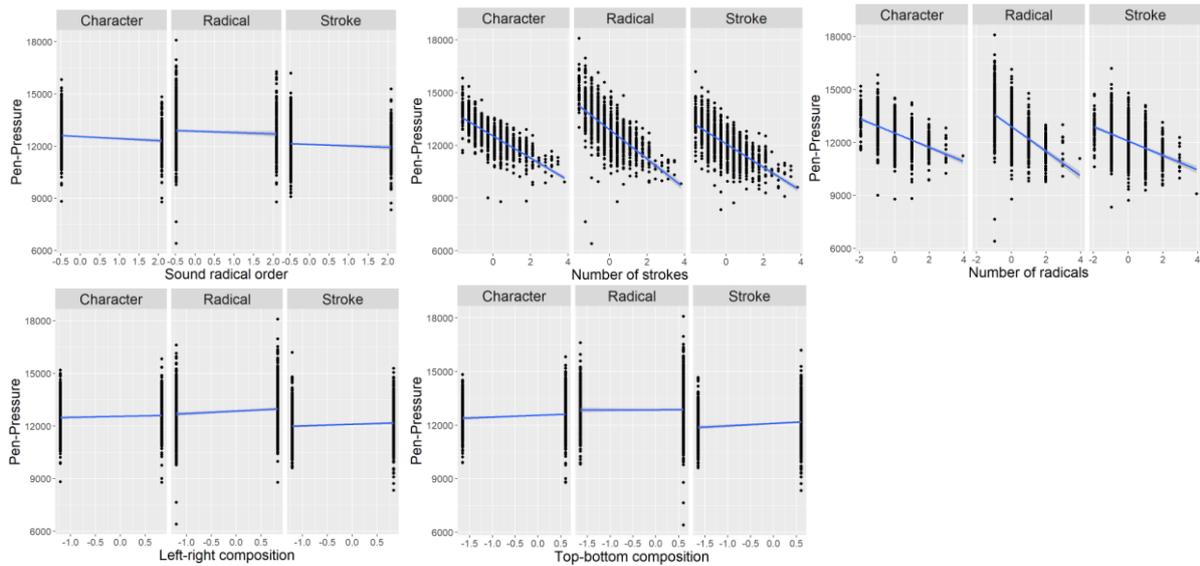

**Figure 9.** Significant interaction effects of pen-pressure as a function of lexical variables, as described in **Table 9**. The lexical variables were transformed into z-scores, include sound radical order, number of strokes, number of radicals, left-right composition, and top-bottom composition.

## DISCUSSION

In this paper, we reported the construction of a large-scale database of character handwriting at different levels (character, radical and stroke). Multiple regression analyses revealed that orthographic predictors significantly affect handwriting at all the three orthographic levels during both handwriting preparation and execution. Phonological factors influence handwriting preparation at the character and radical levels but not at the stroke level; phonological factors impact handwriting execution across all the orthographic levels. Importantly, we found that most of these lexical factors exhibited the strongest lexical effects at the character level, followed by the radical level, with the smallest effects observed at the stroke level. We also introduced an open-source package that we developed to allow researchers with limited programming experience to conduct handwriting studies across various experimental paradigms. Though our setup was designed for writing-to-dictation, it can be easily adapted for other handwriting tasks (e.g., copying). We also provided scripts for batch-processing handwriting data, including temporal, spatial and pen-pressure information at the character, radical, and stroke levels; digitalization of handwritten images, and data visualization for handwriting details.

*The role of phonology in character and sub-character handwriting*

Characters with more transparent phonology-orthography mappings are less susceptible to amnesia. These results replicate the findings by Huang et al. (2021a, 2021b). However, we did not find an effect of character regularity on handwriting latency. This finding contrasts with Wang et al. (2020), who reported that regularity modulated character handwriting latency. The discrepancy can be because Wang et al. (2020) adopted a between-subjects design, in which handwriters were randomly assigned to write only 200 characters selected from a 1600-character cohort, potentially introducing variability across individuals and limiting statistical power to detect regularity effects. In contrast, our within-subjects design required all participants to handwrite 1200 characters via dictation, thereby capturing a more comprehensive range of character properties per individual and controlling for inter-subject variability.

Characters without a sound radical in the first position show lower amnesia rates and shorter writing duration compared with those having a sound radical first. Similarly, phonograms (compared with non-phonograms) are also linked to shorter handwriting execution. We also found that characters with lower homophone density are associated with shorter character writing latency. These results suggest the involvement of phonology-orthography conversion in handwriting, so that handwriters can access a character more easily when it has a sound radical in its conventional position (right or bottom position), when it is a phonogram (which typically displays higher transparency between phonology and orthography compared with non-phonograms), or when it has less competitive homophones. This hypothesis is also supported by the neuroimaging studies: compared with drawing symbols, handwriting Chinese characters elicited greater activation in the brain regions associated with phonological and orthographical processing (e.g., SMG; Xu et al., 2025a; Rapp et al., 2011, 2016; Reich et al., 2003; Lee et al., 2004).

Our study is the first to show that phonological factors modulate writing execution at both the radical and stroke levels. For instance, compared with non-phonograms, phonograms are associated with shorter handwriting duration at the stroke level. Lau (2020a, 2020b) found that phonological predictors influence handwriting latency at the radical level. Our study further shows that phonological effects cascade from radical-level handwriting execution to the stroke level, suggesting that individuals continue to utilize the sub-lexical route to facilitate handwriting at both the radical and stroke levels.

Furthermore, interaction effects indicate that the influence of phonology is larger at the character level than at the radical or stroke level. For instance, our study found that compared with writing latencies at both the radical and stroke levels, character latency has a greater increase for characters with more homophone density. This suggests a hierarchical model of handwriting production, where phonological activation is not uniformly distributed across writing units - The selection and organization of characters are heavily influenced by phonological information, while the execution of the radicals and strokes themselves is less susceptible to this interference. It sheds light on the learning and educational practice, teaching methods that emphasize phonology in character-level orthographic retrieval during handwriting may be more effective than those targeting radical- or stroke-level handwriting.

*The role of semantics in character and sub-character handwriting*

Our study is the first to show that characters with higher concreteness are associated with shorter character writing latency and stroke writing duration, in line with findings that naming latency is shorter for characters or words with higher concreteness (Liu et al., 2007).

We also found that characters with higher imageability are associated with greater radical pen-pressure and these results are in line with findings that naming latency is shorter for characters or words with higher imageability (Cortese, 1997; Liu et al., 2007). These results suggest that concreteness or imageability of characters influences handwriting execution, possibly by facilitating access to sensorimotor simulation (e.g., tactile imagery or visual of the character's meaning for real-world referent) that supports the integration of semantic and motor representations. The semantic modulation appears to cascade from character-level writing preparation to sub-character level fine-grained motor execution. Finally, compared with radical and stroke latencies, character latency showed a greater decrease for characters with higher concreteness. This suggests that concreteness primarily facilitates early-stage lexical access (e.g., character retrieval), while downstream processes (e.g., radical or stroke retrieval) are less affected.

*The role of orthography in character and sub-character handwriting*

We found that higher character frequency is associated with less occurrence of character amnesia, shorter character writing latency, shorter character writing duration, and greater character pen-pressure. These findings align with previous research showing a facilitative effect of lexical frequency on handwriting in both alphabetic languages (e.g., Rapp et al., 2016; Rapp & Dufor, 2011) and Chinese (e.g., Wang et al., 2020; Zhang & Feng, 2017; Huang et al., 2021a, 2021b). Similarly, characters acquired earlier in life are associated with lower character amnesia rates, shorter writing preparation and execution, and greater pen-pressure. More familiar context words are also associated with lower amnesia rates and shorter character writing latency. These results are consistent with previous findings (Wang et al., 2020; Huang et al., 2021a, 2021b), suggesting that characters that are more frequently encountered, learned earlier, or appear in familiar contexts have stronger, more stable representations in the orthographic lexicon, making them easier to retrieve and execute at the character level.

Additionally, our study found that the above-mentioned orthographic factors (e.g., character frequency, age of acquisition, and word familiarity) facilitate writing preparation and execution at the radical or stroke levels. These findings are consistent with Lau (2020a, 2020b), who also found that higher character frequency facilitates sub-character level writing latency. The stability of orthographic representations in the orthographic lexicon is influenced by its frequency (Rapp et al., 2016; Rapp et al., 2011). Our results show that character frequency modulates handwriting at the character and sub-character levels, emphasizing the involvement of orthographic lexicon under the multilevel orthographic–motor integration (e.g., character, radical, and stroke levels). Importantly, these effects are larger at the character level than at the radical level, and larger at the radical level than at the stroke level, suggesting that orthographic lexicon, which contains long-term orthographic representations, continues to influence handwriting at the sub-character level, but its impact diminishes as orthographic representations are decomposed into smaller units.

We also found that characters with fewer strokes are associated with lower character amnesia rates, shorter character writing preparation and execution, and greater pen-pressure. These results are consistent with previous research (Wang et al., 2020; Huang et al., 2021a, 2021b) and challenge findings from small-sample studies suggesting that the number of strokes does not influence orthographic retrieval (Su & Samuels, 2010). Our findings imply that a smaller number of strokes may reduce working memory load, thus allowing individuals to more efficiently program whole-word representations into graphemes (i.e., strokes) before and during handwriting execution. We also found that characters with a left-right composition facilitate character writing preparation and execution, suggesting that the visual–spatial aspect of radical programming influences character-level handwriting. This compositional effect may reflect habitual writing patterns, as writers are generally more accustomed to producing left-right structures than the other configurations.

Additionally, the effects of stroke number and composition cascade to radical and stroke level handwriting preparation and execution. Latency and duration are more strongly influenced by stroke count and composition at the character level, following by radical level, and then stroke level. These findings suggest that the graphemic buffer, a cognitive system that temporarily stores orthographic representations, is continuously engaged during character, radical, and stroke production but diminishes in influence as orthographic representations are decomposed into smaller units. Thus, character-level execution demands the highest orthographic working memory load in order to maintain the complete character representation. Radical level requires less memory load for processing smaller graphemic clusters. Strokes also operate as meaningful units that convey both linguistic and motor information to guide writing execution. These results further suggest that radicals may serve as orthographic chunks in reducing working memory load and as spatial templates that guide stroke execution (Yang et al., 2024). These results are consistent with previous studies demonstrating that participants had shorter handwriting latencies, fewer character amnesia instances when the prime and target shared the same phonetic radical (e.g., Xu et al., 2025b; Zhang & Wang, 2015; 呱 – 狐, "crying" – "fox", sharing the phonetic radical 瓜 "melon") or shared the same semantic radical (Xu et al., 2025b; Damian & Qu 2019; semantic radical 米, meaning "rice", in the character 糕, meaning "cake") than when they did not share any radicals.

Collectively, these findings contribute to refining current handwriting models by highlighting the cascading effects of linguistic components across character, radical, and stroke level writing preparation and execution (**Figure 9**). When hearing a dictation prompt specifying a target Chinese, handwriters first employ the phonological (e.g., the facilitative effects of sound radical order and character regularity in character writing latency; Wang et al., 2020) and semantic information (e.g., the more concrete meaning of the characters facilitate the access of the semantic representations (e.g., Lau et al., 2007) to guide access to the orthographic representations of characters in the orthographic lexicon. More stable orthographic representations (i.e., more frequent characters) are accessed more quickly from orthographic long-term memory during handwriting preparation and execution (e.g., facilitative effects of frequency on character writing latency and duration; Rapp et al., 2016; Rapp et al., 2011). The retrieved orthographic information is maintained in a graphemic buffer; characters with more strokes tend to be produced less accurately and/or more slowly (e.g., Wang et al., 2020; Huang, Zhou, et al., 2021; Huang, Lin, et al., 2021).

At the sub-character level handwriting, the phonological system modulates writing preparation and execution: handwriting durations are shorter for characters with a sound radical to be written later. The semantic system also influences radical-level execution: characters with more imageable meanings elicit greater pen pressure during radical pen-pressure. Finally, graphemic buffer also guides access to the retrieved orthographic information during radical preparation and execution: handwriters are quicker at writing latency and writing duration for characters with fewer strokes and radicals. At the most basic stroke level, stroke preparation still relies on orthographic information retrieved from the orthographic lexicon, this orthographic information is stored in the orthographic working memory for the execution. During stroke execution, both phonological and semantic information re-engage to influence production duration. Finally, the observed interaction between lexical variables and orthographic level, with the writing latency being longest at the character level, followed by the radical level, and shortest at the stroke level, provides empirical support for a three-level orthographic model of handwriting.

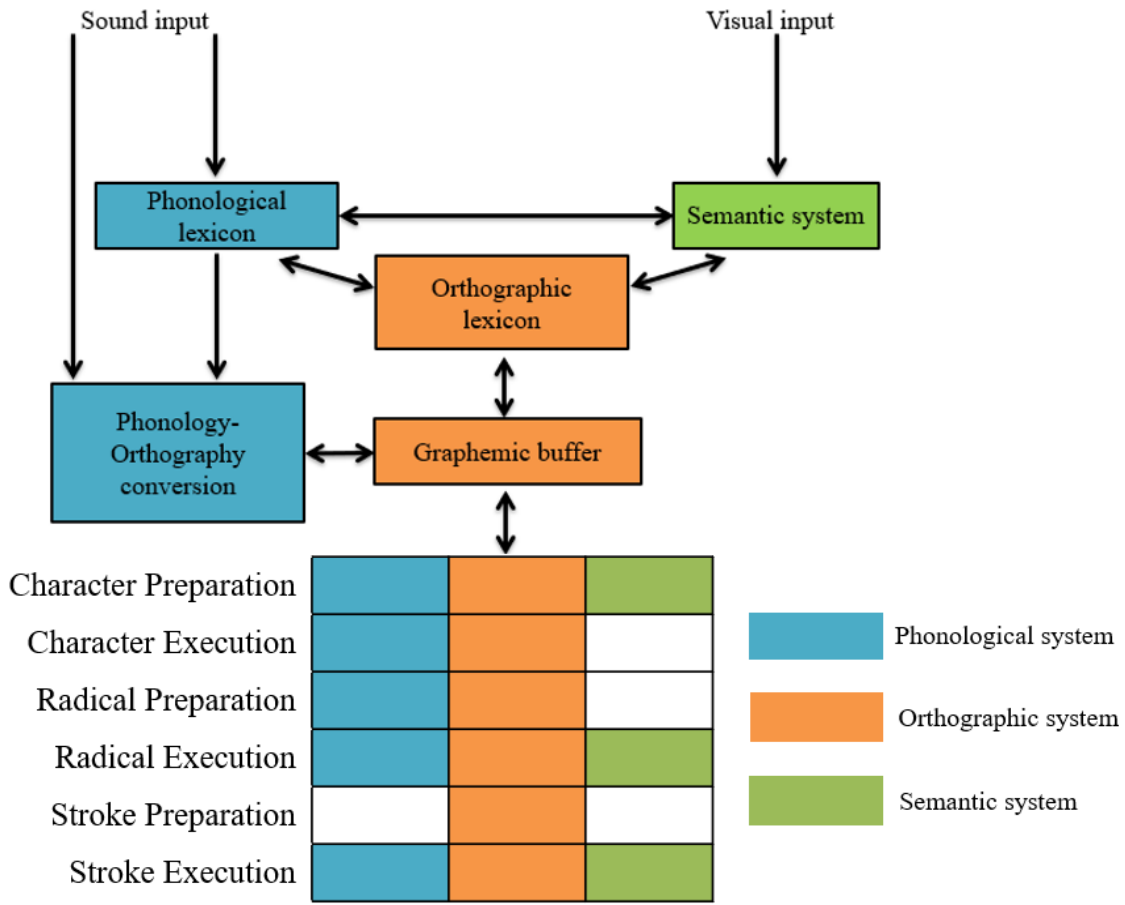

**Figure 9.** An updated cognitive model for handwriting at character, radical, and stroke level handwriting. We present a cascaded model of handwriting where phonological system modulates handwriting preparation and execution at the character and radical levels, it also modulates handwriting execution at the stroke level. Semantic system modulates handwriting preparation at the character level, it also modulates handwriting execution at the radical and stroke levels. Orthographic system has an all-around effect.

*Possible applications of handwriting database and OpenHandWrite_Toolbox*

We expect a wide range of applications for the current handwriting database and toolbox for studying handwriting. This database enables the development of rapid diagnostic tests for assessing orthographic retrieval abilities. In the digital age, reduced handwriting practice has led to the deterioration of explicit orthographic knowledge, resulting in widespread character amnesia. Despite this growing phenomenon, there is currently lack of standardised diagnostic tests available to measure individuals' ability in orthographic retrieval. To address this gap, Langsford et al. (2024) used this dataset to identify 30 characters that are most effective in discriminating adults' character amnesia rates. Their findings demonstrated that the individual' amnesia rate derived from these 30 items exhibited a strong correlation (r ≈ 0.90) with the amnesia rate obtained from a set of 1200 characters. These findings suggest that our database is useful for constructing a concise, 30-item diagnostic test for character amnesia in Chinese.

Recent studies have also employed penscripts (e.g., handwritten images) to evaluate individuals' ability in producing characters with good penmanship (i.e., the ability to handwriting legibly and aesthetically). For instance, the Children's Handwriting Evaluation Scale (CHES; Phelps et al., 1985) relies on trained raters to assess the legibility of children's penscripts based on criteria such as letter forms, slant, spacing, and overall appearance. In the Minnesota Handwriting Assessment (MHA; Reisman, 1999), children copied English words, while occupational therapists evaluated the penscripts on alignment, size, spacing, and form appearance (a similar approach sees also Minnesota Handwriting Test, from Reisman, 1993, and Print Tool from Olsen & Knapton, 2006). However, these subjective assessment tools tend to be costly (e.g., requiring trained raters or teachers for evaluating handwriting) and are prone to individual biases. To address these issues, researchers can utilize the penscripts in this database to develop an automatic assessment system for penmanship. For instance, Xu et al. (2024) made use of the penscripts from a large-scale traditional Chinese handwriting database, combining penmanship ratings for each handwritten samples, they trained a convolutional neural network in predicting penmanship scores. This method achieved a remarkable performance with an overall normalized Mean Absolute Percentage Error of 9.82%, highlighting the potential usage of our database for automated systems of penmanship evaluation in different scripts.

*OpenHandWrite_Toolbox* provides a tool to investigate handwriting in the digital era. Handwriting literacy has been largely declined due to the heavy reliance on digital devices (e.g.,

Almog, 2019). Empirical evidence has indicated that compared with handwriting practice, typing on a computer led to poorer language learning outcomes in terms of spelling (Cunningham & Stanovich, 1990; Reybroeck & Michiels, 2018), letter recognition (Longcamp et al., 2005; Bara & Gentaz, 2011), and handwriting (Chen et al., 2016). Huang, Lin et al. (2021) further demonstrated that college students with greater digital exposure (e.g., frequent use of smartphones or computers) and less engagement with handwriting (e.g., the amount of time spent on handwriting) were more likely to experience character amnesia. Our toolbox can be used to test how digital typing impairs handwriting ability and whether this deterioration occurs mainly at the character, radical, or stroke level. This package can also be used to investigate the neurocognitive processes underlying Chinese character handwriting (see Xu et al. 2025a for more details).

Additionally, this package has the potential in pre-screening children with dysgraphia, a condition characterised by a specific difficulty in acquiring handwriting skills (Berninger et al., 2008; Gosse & Reybroeck, 2020; Kandel et al., 2017; McCloskey & Rapp, 2017). Research have shown that children with dysgraphia exhibited poorer performance in writing characters with lower frequency (e.g., Marinelli et al., 2017; Afonso et al., 2015; Reich et al., 2003), lower regularity (e.g., Brunsdon et al., 2005; Cholewa et al., 2010), or more stroke counts (e.g., Roncoli & Masterson, 2016; Yachini & Friedmann, 2010). Our package can capture the degree of variation in handwriting ability by analysing the features at character, radical, and stroke levels between these two groups. This enables researchers to investigate the cognitive mechanisms underlying developmental dysgraphia and aid in the development of effective diagnostic tools for handwriting difficulties.

**Conclusion**

This study introduces a large-scale database of Chinese character handwriting and demonstrated how lexical variables influence writing latency, duration, and pen-pressure at the character, radical, and stroke levels. Our study highlights the cascading effects of linguistic components on all the three levels of writing preparation and execution. This database was collected by an upgraded package: OpenHandWrite_toolbox, which is a user-friendly and open source toolbox. It includes a GUI for intuitive experiment design and batch-processing scripts to extract varies handwriting measurements. This toolbox is designed for a wider audience to use, adapt, and modify for future handwriting research.

**Appendix A. Guide to experiment building**

OpenHandWrite's GetWrite has been typically used with PsychoPy's scripting interface (https://github.com/isolver/OpenHandWrite/wiki/GetWrite-Experiment-Template). In this case, an experiment needs to be implemented in Python scripts by invoking PsychoPy and GetWrite's functionalities programmatically. While such a programming approach is the most flexible and powerful way to create experiments, it can be non-intuitive and demanding for researchers who are not experienced in Python. To ease the challenges in developing and running handwriting experiments, we provide a fully functional template experiment that can be opened and edited with PsychoPy Builder. A user should launch PsychoPy Builder by running the PsychoPyBuilder.bat file bundled with the OpenHandWrite distribution (downloadable at https://github.com/isolver/OpenHandWrite/releases), which provides built-in access to GetWrite. For a comprehensive introduction and tutorial to PsychoPy Builder, see Peirce and MacAskill (2018).

A conceptual flowchart of the critical procedures of our template experiment is illustrated in **Appendix Figure 1**. At the beginning of the experiment, the program initializes the tablet monitoring interface provided by GetWrite. Then, a pen position validation procedure is run, during which a black circle will appear at nine different locations on the screen, and the participant needs to use the tablet pen to press at a central white point inside the circle each time. After validation, the experiment enters the sequence of experimental trials (see **Apparatus and procedure** for details). Within each trial, the participant's pen movement data is recorded during the writing phase, and various trial variables (e.g., handwriting latency, duration, and self-reports) are saved. Finally, the tablet monitoring interface is closed at the end of the experiment.

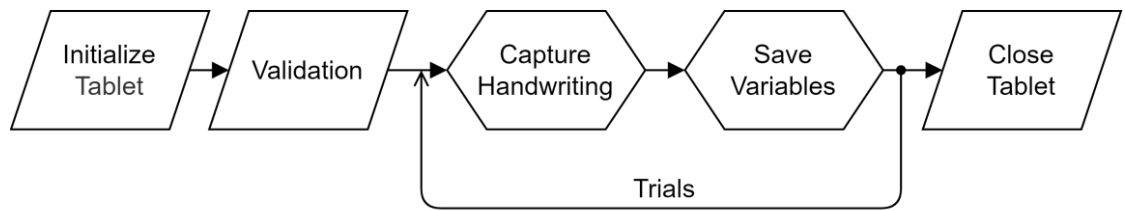

**Appendix Figure 1.** Critical procedures in the PsychoPy experiment.

A user familiar with the PsychoPy Builder GUI would find that most parts of our template are not different from a regular PsychoPy experiment. The overall *flow* of the experiment is constructed by combining various *routines*, which represent the steps of the experiment (e.g., showing instructions, recording handwriting data, collecting self-report responses). A routine consists of one or more *components*, which are the building blocks of an experiment and define the actual behaviour of the routine. For example, the self-report routine contains a text component for displaying the guide text, and a keyboard component for monitoring and collecting the participant's response. A series of routines can be executed repeatedly by grouping them into a *loop*, which is typically used for iterating experimental trials. Trial content (e.g., audio stimulus, target character) and condition information can be conveniently defined in a CSV (comma-separated values) file or Excel sheet (this is called the "condition file"). However, for handwriting experiments, there are a few special requirements regarding the format of the condition file (see **Saving trial variables** below). Because GetWrite is a programming library, there are no GUI components in PsychoPy Builder for accessing GetWrite's functionalities. As a solution, we used PsychoPy Builder's Code component to invoke GetWrite with custom inline codes. In the following sections, we detail the Code components in our template that are critical to capturing handwriting and saving trial variables.

*Initializing and closing the tablet device*

GetWrite uses PsychoPy's ioHub framework for parallel monitoring of pen movements. We use the Code component `init_io` in the `init_seq` routine to start the ioHub-based tablet device (the `tablet` object). Inside this component, the `start_iohub` function begins the ioHub service; the name of the data output file should be passed to the function's first argument. The `data.importConditions` method reads all trial information from the condition file (which is passed to the first argument of the method), and the `io.createTrialHandlerRecordTable` method associates the trial information with the ioHub service. At the end of the experiment, we use the `end_io_code` component in the `end_io` routine to stop the tablet device. At the very beginning of the experiment, we also initialised two visual objects, `pen_pos_stim` and `pen_traces_stim` (inside the `init_wintab` component of the `init` routine), for displaying the latest pen tip location and the existing handwriting traces.

*Capturing handwriting*

Capturing/recording of handwriting is started by setting the `tablet.reporting` property to `True` (`tablet.reporting = True`), and is paused by setting the property to `False` (`tablet.reporting = False`). During each trial, we start recording handwriting before the cue sound (the `main_aud_pen` component of the `main_audio` routine) and stop recording after the participant presses the space bar (the `main_pen` component of the `main_trial` routine). Because participants write characters on paper sheets with a digital ink pen tablet in our experiment, we chose not to draw handwriting traces on the screen to minimize distraction. However, pen traces can be easily shown by uncommenting the following Python code in `main_pen` (under the "Each Frame" tab of the component): `pen_traces_stim.updateFromEvents(pen_samples)`.

*Saving trial variables*

Different from typical PsychoPy experiments, trial variables are saved into an HDF5 file instead of a CSV file because data saving is handled by the ioHub service for GetWrite experiments. Each participant's trial variables, along with the recorded handwriting processes, will be saved to the HDF5 file under the "*data*" folder. In our experiment, the `save_vars_code` component in the `save_vars` routine is responsible for storing participant ID, self-report responses, and several auxiliary variables.

*Condition file format and data saving*

With the current version of OpenHandWrite (v0.4.9), there are three consequences of using the ioHub service for data saving. First, any variable that needs to be saved into the HDF5 file *must* appear in the "condition file". For example, if we want to record participant ID and self-report responses, we must define these variables in the condition file (see the two rightmost columns in **Appendix Figure 2**), even though their values are not determined before the actual experiment is done. Second, the placeholder values for these variables *must* match the data types of their expected resulting values. For example, because self-report responses are expected to be an integer number (0=correct, 1=character amnesia, 2=did not know the correct writing), we used a negative *integer* number (-1) as placeholders. We also used the *character* "x" as the placeholder for participant ID, whose value is expected to be a string of characters. Third, the variables *must* be assigned with their actual values during the experiment (e.g., the variable for self-report must be given the actual participant response codes), and they *must* be saved with the ioHub service (this is done by the `io.addTrialHandlerRecord` method in the `save_vars_code` component).

| | A | B | C | D | E | F | G | H | I | J | K |
|---|---|---|---|---|---|---|---|---|---|---|---|
| 1 | DV_TRIAL_ID | DV_AUD_ONSET | DV_AUD_OFFSET | DV_TRIAL_START | DV_TRIAL_END | ROW_INDEX | audio_name | text | target | participant_id | self_report |
| 2 | -1 | -1.1 | -1.1 | -1.1 | -1.1 | 1 | List1\1ba4.wav | 罢了的罢 | 罢 | x | -1 |
| 3 | -1 | -1.1 | -1.1 | -1.1 | -1.1 | 2 | List1\2ze2.wav | 选择的择 | 择 | x | -1 |
| 4 | -1 | -1.1 | -1.1 | -1.1 | -1.1 | 3 | List1\3diao1.wav | 雕像的雕 | 雕 | x | -1 |
| 5 | -1 | -1.1 | -1.1 | -1.1 | -1.1 | 4 | List1\4guo1.wav | 黑锅的锅 | 锅 | x | -1 |
| 6 | -1 | -1.1 | -1.1 | -1.1 | -1.1 | 5 | List1\5feng3.wav | 讽刺的讽 | 讽 | x | -1 |
| 7 | -1 | -1.1 | -1.1 | -1.1 | -1.1 | 6 | List1\6hai4.wav | 害怕的害 | 害 | x | -1 |
| 8 | -1 | -1.1 | -1.1 | -1.1 | -1.1 | 7 | List1\7fang1.wav | 芳香的芳 | 芳 | x | -1 |

**Appendix Figure 2.** An example condition file.

*Auxiliary variables*

Several important auxiliary variables *must* be defined in the condition file and their actual values also *must* be saved during the experiment. The variable `DV_TRIAL_ID` must correspond to the actual index of the trials (after possible randomization during the experiment). `DV_TRIAL_START` and `DV_TRIAL_END` must store the starting and ending timestamps for the writing phase of each trial; they are critical because MarkWrite will need to extract pen movements according to the timestamps saved in the two variables. Because `DV_TRIAL_START` and `DV_TRIAL_END` are expected to be real numbers (which have decimal places), we used a negative *real* number (-1.1) as their placeholders. Users can define any number of additional variables if needed. For example, we defined `DV_AUD_ONSET` and `DV_AUD_OFFSET` for the onset and offset times of the auditory stimulus; we also stored their actuals values with the `save_vars_code` component.

**Appendix B. Handwriting data segmentation and feature extraction**

The MarkWrite application (v0.4.9) can load the resulting HDF5 file (from GetWrite) that captured the handwriting data and visualise the handwriting data, the Markwrite interface is shown in **Figure 2**. The detected handwriting data includes when the pen in the air is within a pen tablet's capture range (the distance between the pen-tip and PTH-651 tablet is within 3mm), or the pen-tip contact with the tablet to provide the time and coordinates for each pen sample data point. MarkWrite can automatically detect the stroke boundaries based on whether the pen-tip touched the tablet during handwriting. The stroke boundaries are defined by the onset and offset of the stroke. Specifically, the boundary starts when the tablet first detects a pen-pressure sample from the pen-tip, and ends when the tablet registers zero pen-pressure. This auto-segmentation enables us to determine the feature boundaries at stroke, radical or character level. MarkWrite also allows users to manually create feature boundaries, that is, users can determine whether they want to include all the consecutive non-zero pen-pressure samples depending on their research questions (e.g., Chinese writers will sometimes have continuous strokes between two radicals so when auto-segmentation functions not able to recognize there are actually two radicals here, users can choose the boundaries manually). Segments can be nested (e.g., a radical segment can have several 'child' segments for strokes), and they form a segment tree within a specific character. The users can export the pen sample data report plus segments report with the being labelled strokes, radicals or characters.

*Pen sample report & Raw sample data report plus segs report*

There are different types of reports can be exported from the MarkWrite application. We will mainly discuss two export functions that were being used in our R scripts to generate stroke/radical-level database. These reports include all the responses recorded by GetWrite, users can further program with GetWrite's Psychopy interface to define additional variables they want to record (e.g., item number, participant id or self-report). Pen data would be recorded when the pen-tip is close enough that the tablet can detect the pen in the air data (zero pen-pressure) or when the pen-tip touched the tablet (non-zero pen-pressure), including x and y-axis locations, pen-pressure, timestamps, segment labels (if they were additionally defined in MarkWrite by the researcher), and velocity in x or y coordinates.

*Extracting handwriting metrics*

A series of R functions were created for extracting handwriting metrics and are contained in a single R script batch_funcs.R. Inside the script, parallel processing of multiple sessions' data is enabled by R's future package. The calculation of the metrics is based on the Pen Sample Report and Raw Sample Data Report Plus Segs Report, which exposes various information for each pen sample, including the timestamp, x and y coordinates, and pen-pressure. The two Reports can be exported by MarkWrite from the PsychoPy experiment's resulting HDF5 file. Batch export of the Reports can be done with the script run_batchreportgen.bat (which internally calls another script, batchreportgen.py). As detailed below, for each written character in the current database, we generate handwriting metrics at three levels: (1) character, (2) radical (manually segmented), and (3) stroke (see also **Appendix Figure 3**).

At the whole character level, character writing latency (**char_rt**) is equal to duration from offset of the stimulus to the pen-tip firstly touched the tablet on this trial (onset of character handwriting). Character writing duration (**char_dur**) is the duration from the onset of pen-tip firstly touched the tablet on this trial to the last pen sample with non-zero pressure. Character length (**char_len**) is the sum of the lengths of all its strokes belonging to the character, it was calculated by the linear distance between consecutive pen sample points based on the x and y coordinates. The average pen-pressure of the character (**char_press_avg**) is the mean pressure of all stroke samples of the character.

For radical metrics, continuous strokes that straddle two or more radicals (if there are any) will be divided up according to the radical segmentation. Radical writing latency (**radical_rt_rel**) is equal to the duration between offset of the last radical to onset of the current radical. Radical writing duration (**radical_dur**) is the duration between onset of the current radical to the end of the current radical. Radical length (**radical_len**) is the sum of the lengths of the strokes belonging to the radical. The distance to the previous radical (**radical_dist**) is equal to the linear distance between first point of the current radical and last point of previous radical. The average pen-pressure of the radical (**radical_press_avg**) is the mean pressure of the radical's pen samples.

Among these, stroke-level metrics are the most fundamental and serve as the basis for computing higher-level metrics. Stroke writing latency (**stroke_rt_rel**) is calculated by subtracting the last stroke's offset time from the timestamp of the current stroke writing onset time. Stroke writing duration (**stroke_dur**) is the difference between the timestamps of the last and first samples. Stroke length (**stroke_len**) is calculated by summing up the linear distance between consecutive pen sample points (based on their x and y coordinates), which is analogue to connecting the samples with straight lines. The raw length is converted to millimeters by dividing it with the tablet's lines per millimeter (lpmm) parameter. The lpmm value is the spatial resolution of a tablet. This value can usually be found in a tablet's specification sheet and can be calculated manually: draw vertical and horizonal lines that cross the whole surface of the tablet, export a pen sample report from MarkWrite and find the maximal y or x coordinates, divide them by physical height or width of the tablet (in millimeters) and the result is the lpmm value. The distance to the previous stroke (**stroke_dist**) is the linear distance between the current stroke's first sample and the previous stroke's last sample. The average pen-pressure of the stroke (**stroke_press_avg**) is the mean pressure of the pen samples in the stroke.

| Subject | DV_TRIAL_ID | ROW_INDEX | self_report | target | char_dur | char_rt | char_len | char_press_avg | rad_label | rad_dur | rad_rt_rel | rad_len | rad_dist | rad_press_avg | stroke_label | stroke_dur | stroke_rt_rel | stroke_len | stroke_dist | stroke_press_avg |
|---|---|---|---|---|---|---|---|---|---|---|---|---|---|---|---|---|---|---|---|---|
| 1 | 192 | 10 | 0 | 稻 | 2520 | 1369 | 42.46 | 16728 | 1 | 849 | NA | 17.89 | NA | 16680 | 3 | 135 | 72 | 6.9 | 1.195 | 16749 |
| 1 | 192 | 10 | 0 | 稻 | 2520 | 1369 | 42.46 | 16728 | 1 | 849 | NA | 17.89 | NA | 16680 | 4 | 81 | 97 | 2.645 | 4.29 | 14948 |
| 1 | 192 | 10 | 0 | 稻 | 2520 | 1369 | 42.46 | 16728 | 1 | 849 | NA | 17.89 | NA | 16680 | 5 | 75 | 71 | 2.065 | 1.625 | 15049 |
| 1 | 192 | 10 | 0 | 稻 | 2520 | 1369 | 42.46 | 16728 | 2 | 544 | 123 | 7.245 | 5.76 | 14685 | 6 | 106 | 123 | 3.53 | 5.76 | 18220 |
| 1 | 192 | 10 | 0 | 稻 | 2520 | 1369 | 42.46 | 16728 | 2 | 544 | 123 | 7.245 | 5.76 | 14685 | 7 | 60 | 88 | 0.9 | 1.525 | 12859 |
| 1 | 192 | 10 | 0 | 稻 | 2520 | 1369 | 42.46 | 16728 | 2 | 544 | 123 | 7.245 | 5.76 | 14685 | 8 | 66 | 78 | 0.915 | 1.07 | 9191 |
| 1 | 192 | 10 | 0 | 稻 | 2520 | 1369 | 42.46 | 16728 | 2 | 544 | 123 | 7.245 | 5.76 | 14685 | 9 | 67 | 79 | 1.9 | 1.555 | 16521 |
| 1 | 192 | 10 | 0 | 稻 | 2520 | 1369 | 42.46 | 16728 | 3 | 924 | 80 | 17.33 | 1.535 | 17815 | 10 | 188 | 80 | 4.045 | 1.535 | 17949 |
| 1 | 192 | 10 | 0 | 稻 | 2520 | 1369 | 42.46 | 16728 | 3 | 924 | 80 | 17.33 | 1.535 | 17815 | 11 | 165 | 111 | 5.215 | 4.38 | 21782 |
| 1 | 192 | 10 | 0 | 稻 | 2520 | 1369 | 42.46 | 16728 | 3 | 924 | 80 | 17.33 | 1.535 | 17815 | 12 | 53 | 110 | 1.13 | 4.165 | 12311 |
| 1 | 192 | 10 | 0 | 稻 | 2520 | 1369 | 42.46 | 16728 | 3 | 924 | 80 | 17.33 | 1.535 | 17815 | 13 | 211 | 86 | 6.94 | 1.095 | 16065 |

**Appendix Figure 3.** Example of a handwritten trial from the. Subject: number of the subject; DV_TRIAL_ID: number of trial; ROW_INDEX; item index; self-report: participant's report on handwriting, 0, 1, and 2 represent correct handwriting, character amnesia, and incorrect handwriting; target: the to be written character; char_dur: character writing duration; char_rt: character writing latency; char_len: character length; char_press_avg: the average pen-pressure of the character; radical_label: radical index; rad_dur: radical writing duration; rad_rt_rel: radical writing latency; rad_len: radical length; rad_dist: the distance to the previous radical; rad_press_avg: the average pen-pressure of the radical; stroke_label: stroke index; stroke_dur: stroke writing duration; stroke_rt_rel: stroke writing latency; stroke_len: stroke length stroke_dist: the distance to the previous stroke; stroke_press_ave: the average pen-pressure of the stroke.

In addition to extracting the numeric writing metrics above, our R script also features rich visualization functionalities. It can automatically generate the plot for each written character in the *plots\char output* folder (**Figure 3**) and generate detailed stroke-by-stroke plots in the *plots\by-stroke* folder (**Figure 4**). To batch process the results of multiple experimental sessions, one can first organize the Pen Sample Reports and Raw Sample Data Report Plus Segs Reports exported from the HDF5 files into a single folder, and then write a R script to invoke the process_all_reports function defined in the batch_funcs.R script. Our script batch_summarise_all.R is a complete example for batch processing.